\documentclass{article}

\usepackage[preprint]{neurips_2022}

\usepackage[utf8]{inputenc} 
\usepackage[T1]{fontenc}    
\usepackage{hyperref}       
\usepackage{url}            
\usepackage{booktabs}       
\usepackage{amsfonts}       
\usepackage{nicefrac}       
\usepackage{microtype}      
\usepackage{xcolor}         
\usepackage{amsmath}
\DeclareMathOperator*{\argmin}{\arg\!\min}
\usepackage{amssymb}
\usepackage[title]{appendix}
\usepackage[ruled,vlined]{algorithm2e}
\usepackage{multicol}
\usepackage{multirow}
\usepackage[pdftex]{graphicx}
\usepackage{float} 
\usepackage{adjustbox}

\usepackage{natbib}
\renewcommand{\arraystretch}{2.0}

\usepackage{amsthm}
\usepackage{apxproof}
\newtheorem{assumption}{Assumption}
\newtheoremrep{theorem}{Theorem}
\newtheoremrep{corollary}{Corollary}[theorem]
\newtheoremrep{lemma}{Lemma}
\newtheorem{remark}{Remark}

\title{ Federated Hypergradient Descent }

\author{%
  Andrew K Kan \\
  Naval Information Warfare Center Pacific \\
  \texttt{andrew.k.kan.civ@us.navy.mil} \\
  \texttt{andrew.k.kan@gmail.com} \\
}

\begin{document}

\maketitle

\begin{abstract}
    In this work, we explore combining automatic hyperparameter tuning and optimization for federated learning (FL) in an online, one-shot procedure.  We apply a principled approach on a method for adaptive client learning rate, number of local steps, and batch size.  In our federated learning applications, our primary motivations are minimizing communication budget as well as local computational resources in the training pipeline.  Conventionally, hyperparameter tuning methods involve at least some degree of trial-and-error, which is known to be sample inefficient.  In order to address our motivations, we propose FATHOM (Federated AuTomatic Hyperparameter OptiMization) as a one-shot online procedure.  We investigate the challenges and solutions of deriving analytical gradients with respect to the hyperparameters of interest.  Our approach is inspired by the fact that, with the exception of local data, we have full knowledge of all components involved in our training process, and this fact can be exploited in our algorithm impactfully.  We show that FATHOM is more communication efficient than Federated Averaging (FedAvg) with optimized, static valued hyperparameters, and is also more computationally efficient overall.  As a communication efficient, one-shot online procedure, FATHOM solves the bottleneck of costly communication and limited local computation, by eliminating a potentially wasteful tuning process, and by optimizing the hyperparamters adaptively throughout the training procedure without trial-and-error.  We show our numerical results through extensive empirical experiments with the Federated EMNIST-62 (FEMNIST) and Federated Stack Overflow (FSO) datasets, using FedJAX as our baseline framework.

\end{abstract}

\section{Introduction}
Federated learning (FL) for on-device applications has its obvious social implications, due to its inherent privacy-protection feature.  It opens up a broad range of opportunities to allow a massive number of devices to collaborate in developing a shared model by retaining private data on the devices.  The ubiquity of machine learning (ML) on consumer data, coupled with the growth of privacy concerns, has pushed researchers and developers to look for new ways to protect and benefit end-users.  In order for FL to deliver its promise in deployed applications, there are still many open challenges remained to be solved.  We are especially interested in the overall communication efficiency of the FL pipeline for it to be realistically deployed in a unique communication environment over expensive links.  To begin, consider a typical step in a machine learning (ML) pipeline: hyperparameter tuning.   Whether it is in a centralized, distributed or federated setting, it is an essential step to achieve an optimal operation for the training process.  At the heart of an ML training process is the optimization algorithm.  In particular, we are interested in using Federated Averaging (FedAvg) as our baseline federated optimization algorithm for our work.  This is because, despite all the recent innovations in FL since its introduction in 2016 by \citet{mcmahan2016communicationefficient}, FedAvg remains the de facto standard in federated optimization for both research and practice, due to its simplicity and empirical effectiveness.  In order for FedAvg to operate effectively, it requires properly tuned hyperparameter values.  

Our work focuses specifically on hyperparameter optimization (HPO) of: 1) client learning rate, 2) number of local steps, as well as 3) batch size, for FedAvg.  We propose FATHOM (Federated AuTomatic Hyperparameter OptiMization), which is an online algorithm that operates as a one-shot procedure.  
In the rest of this paper, we will go through a few notable recent state-of-the-art works on this topic, and make justifications for our new approach.  Then we will derive a few key steps for our algorithm, followed by a theoretical convergence bound for adaptive learning rate and number of local steps in the non-convex regime.  Lastly, we present numerical results on our empirical experiments with neural networks on the FEMNIST and FSO datasets.

Our contributions are as follows:
\begin{itemize}
\item We derive gradients with respect to client learning rate and number of local steps for FedAvg, for an online optimization procedure.  We propose FATHOM, a practical one-shot procedure for joint-optimization of hyperparameters and model parameters, for FedAvg. 
\item We derive a new convergence upper-bound with a relaxed condition (see Section \ref{sec:theory} and remark \ref{rem:relax}), to highlight the benefits from the extra degree-of-freedom that FATHOM delivers for performance gains.
\item We present empirical results that show state-of-the-art performance.  To our knowledge, we are the first to show gain from an online HPO procedure over a well-tuned equivalent procedure with fixed hyperparameter values.
\end{itemize}

\section{Related Work and Justifications for FATHOM} \label{sec:related}
We explore the question whether the FATHOM approach is justified over the more recent, state-of-the-art methods that are designed for the same goal: a single-shot online hyperparamter optimization procedure for FL. \citet{zhou2022singleshot} proposed Federated Loss SuRface Aggregation (FLoRA), a general single-shot HPO for FL, which works by treating HPO as a black-box problem and by performing loss surface aggregation for training the global model. \citet{khodak2021federated} draws inspiration from weight-sharing in Neural Architectural Search (\citet{pmlr-v80-pham18a}, \citet{cai2018proxylessnas}), and proposed FedEx, which is an online hyperparameter tuning algorithm that uses exponentiated gradients to update hyperparameters.  On the other hand, \citet{mostafa2019robust}'s RMAH and \citet{guo2022auto}'s Auto-FedRL both use REINFORCE (\citet{10.1007/BF00992696}) in their agents to update hyperparameters in an online manner, by using relative loss as their trial rewards.   One basic assumption among these methods, is that at least some of the gradients with respect to the hyperparameters are unavailable directly.   Generalized techniques are used to update these quantities, involving Monte-Carlo sampling and evaluation with held-out data.  One key benefit with techniques such as these is their generalizability for a wide range of different hyperparameters.  On the other hand, we identify a few areas with these methods that we would like to improve on.  One, information about the internals of the procedure can and should be exploited.  Two, communication overhead becomes a concern, since sufficient Monte-Carlo sampling is required for some of these techniques to converge, an example being the re-parametrization trick (\citet{kingma2013autoencoding}) which is used for FedEx, RMAH and Auto-FedRL.  From initial observations of their empirical results, while these methods are successful in hyperparameter tuning and reaching target model accuracy as shown in these works, these goals are achieved in unspecified numbers of total communication rounds from works based on RL approaches such as \citet{mostafa2019robust} and \cite{guo2022auto}.

The above observations justify exploring our problem differently from previous approaches.  Our method exploits full knowledge of the training process, and it does not require sufficient trials at potential expense of communication budget.  Inspired by the hypergradient descent techniques developed by \citet{baydin2017online} and \citet{amid2022stepsize} for centralized optimization learning rate, we develop FATHOM by directing deriving analytical gradients with respect to the hyperparameters of interest.  The result is a sample efficient method which offers both improvements in communication efficiency and reduced local computation in a single-shot online optimization procedure.  Meanwhile, FATHOM is not as flexibly applicable in optimizing a wide range of hyperparameters, since each gradient needs to be derived separately to take advantage of our full knowledge of the training process.  We believe this approach is a performance advantage, at the expense of its flexibility. 

There are other notable relevant works.  \citet{charles2020outsized} and \citet{li2019convergence} proved that reducing the client learning rate during training is necessary to reach the true objective.  Yet, a line of interesting works, such as \citet{dai2020federated} and \citet{holly2021evaluation}) applies Bayesian Optimization (BO) on federated hyperparamter tuning, by treating it as a closed-box optimization problem. \citet{dai2021differentially} further updates their use of BO in FL by incorporating differential privacy. However, these BO-based works do not consider adaptive hyperparameters.  Yet, another work (\citet{wang2018adaptive}) shares similarity to our approach of optimally adapting the number of local steps, with their adaptive communication strategy, AdaComm, in the distributed setting.  However, their main interest is reducing wall-clock time.  Lastly, around the same time of this writing, \citet{wang2022fedhpob} publishes their benchmark suite for FL HPO, called FedHPO-B, which would be valuable to our future work.

\section{Methodology} \label{sec:methodology}

\begin{toappendix}
This supplementary section contains missing proofs from Section \ref{sec:methodology}.
\end{toappendix}

In this section we formalize the problem of hyperparameter optimization (HPO) for FL.
We first review FedAvg, a de facto standard of federated optimization methods for research baseline and practice.  Then, we present our method for online-tuning of its hyperparameters, specifically client learning rate, number of local steps, and batch size.  We call our method FATHOM (Federated AuTomatic Hyperparameter OptiMization).

\subsection{Problem Definition}

In this paper, we consider the empirical risk minimization (ERM) across all the client data, as an unconstrained optimization problem:
\begin{equation} \label{eq:erm}
    f^{*} := \min_{x \in \mathbb{R}^{d}} \Bigg[f(x) := \frac{1}{m} \sum_{i=1}^{m} f_{i}(x) \Bigg]
\end{equation} 
where $f_{i}:\mathbb{R}^{d} \rightarrow \mathbb{R}$ is the loss function for data stored in local client index $i$ with $d$ being the dimension of the parameters $x$, $m$ is number of clients, and $f^{*} = f(x_{*})$ where $x_{*}$ is a stationary solution to the ERM problem in eq(\ref{eq:erm}).  

To facilitate some of the discussions that follow, it helps to define assumptions here as we do throughout the rest of this paper:
\begin{assumption} \label{assump:unbiased}
(Unbiased Local Gradient Estimator) Let $g_{i}(x)$ be the unbiased, local gradient estimator of $\nabla f_{i}(x)$, i.e., $\mathbb{E}[g_{i}(x)] = \nabla f_{i}(x)$, $\forall x$, and $i \in [m]$.
\end{assumption}

\subsection{Federated Optimization and Tuning of Hyperparameters}

\paragraph {Federated Averaging (FedAvg)} We describe the operations of FedAvg from \citet{mcmahan2016communicationefficient}, as follows.  At any round $t$, each of the $m$ clients takes a total of $K_{i}$ local SGD steps, where $K_{i} = \lfloor E \nu_{i} / B \rfloor$, and where $\nu_{i}$ is the number of data samples from client index $i$, $B$ is batch size, with epoch number $E=1$ being a common baseline. In this version of FedAvg, heterogeneous data size is accommodated across clients, and the number of local steps can be manipulated via $E$ and $B$ as hyperparameters.  Each local SGD step updates the local model parameters of each client $i$ as follows: $x^{i}_{t, k+1} = x^{i}_{t, k} - \eta_{L} g_{i}(x^{i}_{t, k})$, where $\eta_{L}$ is the local learning rate and $k \in [K]$ is the local step index.  To conclude each round, these clients return the local parameters $x^{i}_{t, K_{i}}$ to the server where it updates its global model, with $x_{t+1} = \sum_{i} \nu_{i} x^{i}_{t, K} / \nu$ where $\nu = \sum_{i}\nu_{i}$.  To facilitate some of the discussions that follow, we define the following quantities:
\begin{equation} \label{eq:deltabar}
    \overline{\Delta}_{t}  \triangleq x_{t+1} - x_{t} = \sum_{i=1}^{m} \frac{\nu_{i}}{\nu} \Delta_{t}^{i}  \quad \text{where} \quad
    \Delta_{t}^{i}  \triangleq - \sum_{k=0}^{K_{i}-1} \eta_{L,t} g_{i}(x_{t}^{i,k}) 
\end{equation}

\paragraph{Offline Hyperparameter Tuning}
Offline tuning is best to be summarized as follows.  We first define $U = \{u \in \mathbb{R} \mid u \geq 0 \}$ with $\eta_{L} \in U$, and $V = \{v \in \mathcal{I} \mid v \geq 1\}$ with $K \in V$.  We also define $C = U \times V$, and $c = (\eta_{L}, K)$, where $c \in C$.  Offline tuning would have the following objective: $\ \min_{c \in C} f_\text{valid}(x, c) \ \  \text{s.t.} \ \ x = \argmin_{z \in \mathbb{R}^{d}} f_\text{train}(z, c) \ $.  With abuse of notation, we use $f_\text{valid}$ for the objective function calculated from a validation dataset which is usually held-out before the procedure, and $f_\text{train}$ for the objective from training data which usually is just local client data.  A few notable offline tuning methods are as follows.  Global grid-search from \citet{holly2021evaluation} is an example of offline tuning that iterates over the entire search grid defined as $C$, completing an optimization process for each grid point and evaluating the result with a held-out validation set.  Global Bayesian Optimization from \citet{holly2021evaluation} is another similar example of offline tuning that follows the same template and objective.  Instead of brute-force grid-search, $c$ is sampled from a distribution $\mathcal{D}_{C}$ over $C$, i.e. $c \sim \mathcal{D}_{C}$, that updates after every iteration.

\paragraph {Online Hyperparameter Optimization}
We are interested in an online procedure that combines hyperparameter optimization and model parameter optimization, with the following objective:
\begin{equation} \label{eq:onlineobj}
    \min_{\substack{x \in \mathbb{R}^{d} \\ c \in C}} f_\text{train}(x, c)
\end{equation}
This formulation is the objective of our method, FATHOM, which we will discuss shortly in detail.  It has the advantage of joint optimization in a one-shot procedure. Furthermore, it does not assume the availability of a validation dataset.  

\subsection{Our Method: FATHOM}
In this section we will introduce our method, FATHOM (Federated AuTomatic Hyperparameter OptiMization).  Recall from our joint objective, eq(\ref{eq:onlineobj}), that both the model parameters, $x$, and hyperparameters of the optimization algorithm, $c$, are optimized jointly to minimize our objective function.  An alternative view is to treat $c$ as part of the parameters being optimized in a classic formulation, i.e. $min_{y}f(y)$ with $y = (x, c)$. As previously mentioned, our method is inspired by hypergradient descent from \citet{baydin2017online} and by exponentiated gradient from \citet{amid2022stepsize}, both proposed for centralized learning rate optimization.  We will present how FATHOM exploits our knowledge of analytical gradients to update client learning rate, number of local steps, as well as batch size, for an online, one-shot optimization procedure.

\begin{assumption} \label{assump:convex}
(Convexity w.r.t. $\eta_{L}$ and $K$) 
We assume $\mathbb{E}_{t}(f(x_{t}))$ is convex w.r.t. $\eta_{L}$ and $K$, even though we assume non-convexity w.r.t. $x_{t}$).  Specifically, convexity w.r.t. $K$ follows the definition in \citet{Murota1998DiscreteCA}, to accommodate the integer space where $K$ is defined.
\end{assumption}

\begin{remark}
Assumption \ref{assump:convex} is necessary to guarantee the existence of subgradients derived in Theorems \ref{thm:piecew} and \ref{thm:piecew2}, and it will be assumed for this work.  In problems dealing with deep neural networks, it is reasonable to not assume convexity w.r.t. hyperparameters.  However, from our empirical results, we claim that the proposed algorithm is still able to operate as desired under this condition. 
\end{remark}

\subsubsection{Hypergradient for Client Learning Rate}
In this section, we derive the hypergradient for client learning rate in a similar fashion as \citet{baydin2017online}, with the difference being that they are mainly concerned with the centralized optimization problem, and that we are concerned with the distributed setting where clients take local steps.  We derive the following hypergradient of the objective function as defined in eq(\ref{eq:erm}), taken with respect to the learning rate $\eta_{L, t-1}$ such that it can be updated to obtain $\eta_{L, t}$:
\begin{align}
    H_{t} & = \frac{\partial f(x_{t})}{\partial \eta_{L, t-1}} = \frac{\partial f(x_{t})}{\partial x_{t}} \cdot \frac{\partial (x_{t-1} + \overline{\Delta}_{t-1})}{\partial \eta_{L, t-1}} = \nabla f(x_{t}) \cdot  \frac{\partial \overline{\Delta}_{t-1}}{\partial \eta_{L, t-1}} \label{eq:hyp1}
\end{align}
where $\overline{\Delta}_{t}$ is the update step for the global parameters $x_t$ as defined in eq(\ref{eq:deltabar}), leading to $\frac{\partial\overline{\Delta}_{t}}{\partial\eta_{L,t}} = \frac{\overline{\Delta}_{t}}{\eta_{L,t}} = - \sum_{i=1}^{m} \frac{\nu_{i}}{\nu} \sum_{k=0}^{K-1} g_{i}(x_{t}^{i,k})$.  We also make the approximation $x_{t+1} - x_{t} = \overline{\Delta}_{t} \approx -\eta_{L, t} \nabla f(x_{t})$.  We can then write the normalized update, $\overline{H}_{t}$, similar to \citet{amid2022stepsize}, as follows:
\begin{align}
    \overline{H}_{t} & = \frac{\nabla f(x_{t})}{\|\nabla f(x_{t})\|} \cdot  \Big(\frac{\partial \overline{\Delta}_{t-1}}{\partial \eta_{L, t-1}} \Big/ \Big\|\frac{\partial \overline{\Delta}_{t-1}}{\partial \eta_{L, t-1}}\Big\|\Big)  \approx -\frac{\overline{\Delta}_{t}}{\|\overline{\Delta}_{t}\|} \cdot \frac{\overline{\Delta}_{t-1}}{\|\overline{\Delta}_{t-1}\|} \label{eq:hyp1approx}
\end{align}
The resulting hypergradient is a scalar, as expected, and can be used efficiently as part of the update rule for $\eta_{L}$, which we will see in Section \ref{sec:updates}.  The implementation is communication efficient, since in each round, each client needs one extra scalar to send back to the server, and likewise the server needs to broadcast one extra scalar back to the clients.  It is also computationally efficient since it avoids calculating the full local gradient $\nabla f(x_{t})$.

\subsubsection{Hypergradient for Number of Local Steps}
Since the number of local steps is an integer, i.e. $K = \{ k \in \mathbb{I} \mid k \geq 1 \}$, this means $f(x_{t})$ does not exist for non-integer values of $K$.  We formulate a subgradient as a surrogate of the hypergradient $\partial f(x_{t}) / \partial K$, as follows.  We will call this a hyper-subgradient.   
\begin{theoremrep} \label{thm:piecew}
When a piecewise function $L_{t}$ is defined for every value of $K_{0} \in [K]$ on $l$, such that $0.0 \leq l < 1.0$, we claim, under Assumption \ref{assump:convex}, that the following is a subgradient of $f(x_{t})$ at $K_{t} = K_{0}$:
\begin{equation} \label{eq:piecew}
    \frac{\partial L_{t}}{\partial l}  = \nabla f(x_{t}) \cdot \big( -\eta_{L,t} \sum_{i=1}^{m} g_{i}(x_{t-1}^{i,K_{t}-1})\frac{\nu_{i}}{\nu} \big)
\end{equation}
where $l$ represents the marginal fraction of local steps beyond $K_{0}$.  We leave the proof (with an illustration in Figure \ref{fig:figproof}) in the Appendix section beginning in eq(\ref{eq:piecew-L}).
\end{theoremrep}
\begin{proof}
We clarify this quantity by re-writing $L_{K_{0}}(l) = L_{t}$ and its definition (with some abuse of notation):
\begin{equation} \label{eq:piecew-L}
    L_{K_{0}}(l) = f(x_{t}) \bigg|_{K = K_{0} + l} = f\Big(x_{t-1} - \eta_{L,t} \sum_{i=1}^{m} \frac{\nu_{i}}{\nu} \sum_{k=0}^{K_{0}-2} \big(g_{i}(x_{t-1}^{i,k}) + l g_{i}(x_{t-1}^{i, K_{0}-1})\big)\Big)  
\end{equation}
for $K_{0} \geq 2$, and $L_{K_{0}}(l) = f\big(x_{t-1} - \eta_{L,t} \sum_{i=1}^{m}  l g_{i}(x_{t-1}^{i, k=0}) \frac{\nu_{i}}{\nu} \big) $ for $K_{0} = 1$, where $l$ represents the marginal fraction of local steps $K_{0}$.  Then, by convexity from Assumption \ref{assump:convex}, and by recalling that $l$ represents the marginal fraction of local steps, we claim that:
\begin{equation}
    \frac{\partial L_{K_{0}}(l)}{\partial l} \leq \Big[ \frac{f(x_{t}) \big|_{K = K_{0} + l} - f(x_{t})\big|_{K = K_{0}}}{l} \Big]
\end{equation}
From this result, we conclude that $\frac{\partial L_{K_{0}}(l)}{\partial l}$ is a subgradient of $f(x_{t})$ at $K = K_{0}$.
With abuse of notation by setting $K_{0} = K_{t}$ and therefore $L_{t} = L_{K_{0}}(l)$, we further derive $\frac{\partial L_{t}}{\partial l}$ by breaking it down similarly as eq(\ref{eq:hyp1}), leads us to:
\begin{align}
    \frac{\partial L_{t}}{\partial l} & = \nabla f(x_{t}) \cdot \frac{\partial \big(- \eta_{L,t} \sum_{i=1}^{m} \frac{\nu_{i}}{\nu} \sum_{k=0}^{K_{t}-2} l g_{i}(x_{t-1}^{i, K_{t}-1})\big)}{\partial l} \\
    & = \nabla f(x_{t}) \cdot \big( -\frac{\eta_{L,t}}{m} \sum_{i=1}^{m} \label{eq:hyp2} g_{i}(x_{t-1}^{i,K_{t}-1}) \big)
\end{align}
The above result concludes the proof for Theorem \ref{thm:piecew}.

\begin{figure} [H]
  \centering
  \makebox[0pt]{\includegraphics[scale=0.29]{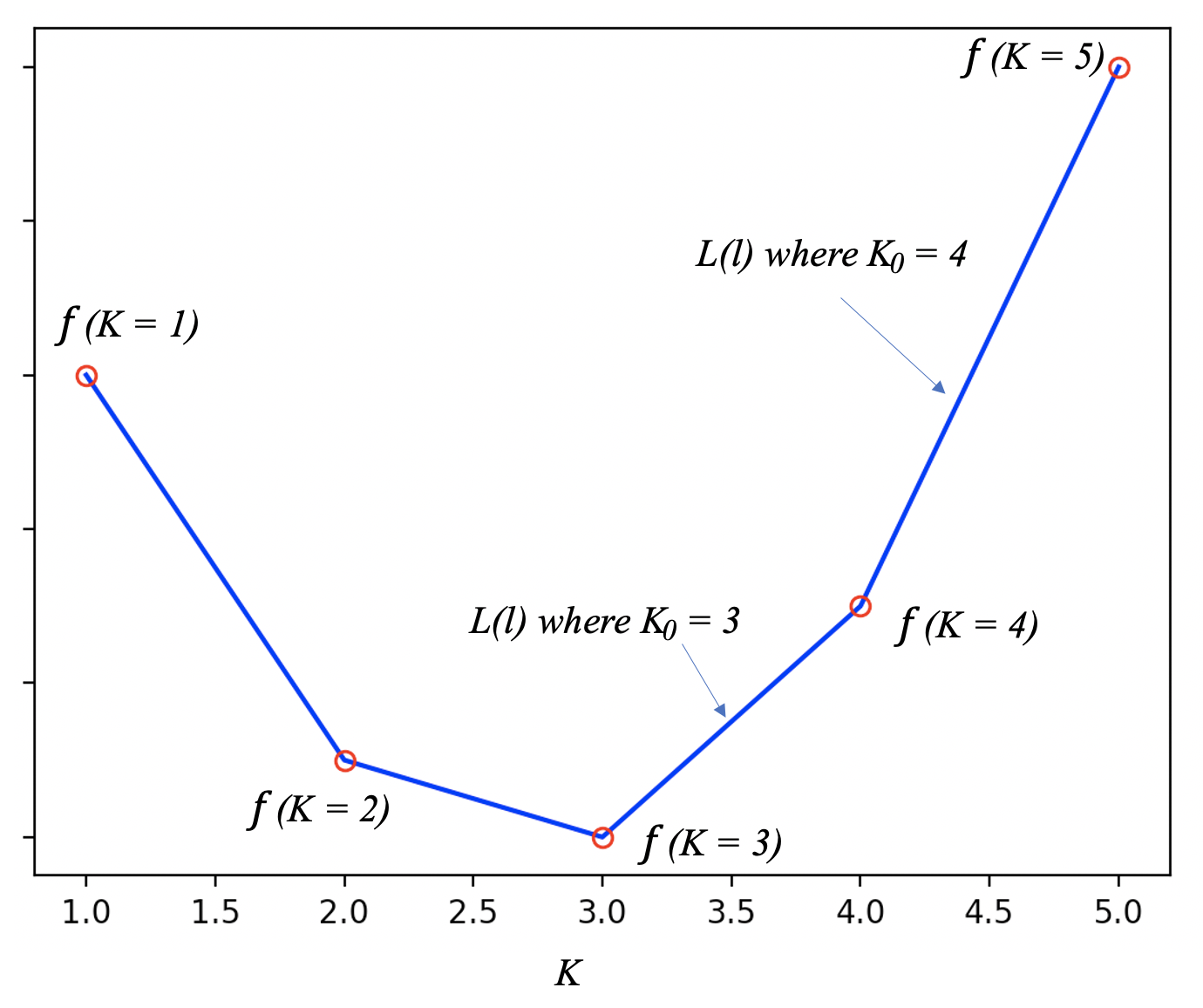}} 
  \caption{ Illustration of piecewise function $L_{K_{0}}(l)$ and $f(K)$, where $f(K) = f(x, c)$ from eq \ref{eq:onlineobj}, and where we ignore $\eta_{L}$ and $x_t$ for this discussion. Notice $K$ is only defined on $\{\mathcal{I} \mid K \geq 1\}$.  Hence $f(K)$ are illustrated as red dots.  However, since $0.0 \leq l < 1.0$, $L_{K_{0}}(l)$ extends from $f(K_0)$ to $f(K_0 + 1)$. } \label{fig:figproof}
\end{figure}
\end{proof}

The result from Theorem \ref{thm:piecew} is not sufficiently communication-efficient for implementing an update rule for $K$.  This is because it would require the quantity $g_{i}(x_{t-1}^{i,K_{t}-1})$ to be communicated from each client $i$ to the server.  To save communication, let us reuse what the server has in memory: $\overline{\Delta}_{t} = \big( -\eta_{L} \sum_{i=1}^{m} \frac{\nu_{i}}{\nu} \sum_{k=0}^{K_{t}-1} g_{i}(x_{t}^{i,k}) \big)$.  If we let:
\begin{align}
    S_{t} & = \nabla f(x_{t}) \cdot \big( -\eta_{L,t} \sum_{i=1}^{m} \frac{\nu_{i}}{\nu} \sum_{k=0}^{K_{t}-1} g_{i}(x_{t-1}^{i,k}) \big) l \\
    N_{t} = \frac{\partial S_{t}}{\partial l} & = \nabla f(x_{t}) \cdot \big( -\eta_{L,t} \sum_{i=1}^{m} \frac{\nu_{i}}{\nu} \sum_{k=0}^{K_{t}-1} g_{i}(x_{t-1}^{i,k}) \big)
    = \nabla f(x_{t})  \cdot  \overline{\Delta}_{t-1} \label{eq:piecew-mod} \\
    \overline{N}_{t} & = \frac{\nabla f(x_{t})}{\|\nabla f(x_{t})\|} \cdot \frac{\overline{\Delta}_{t-1}}{\|\overline{\Delta}_{t-1}\|} \approx -\frac{\overline{\Delta}_{t}}{\|\overline{\Delta}_{t}\|} \cdot \frac{\overline{\Delta}_{t-1}}{\|\overline{\Delta}_{t-1}\|} \label{eq:piecew-modapprox}
\end{align}
where eq(\ref{eq:piecew-modapprox}) is the normalized update as in \citet{amid2022stepsize}.  We claim that eq(\ref{eq:piecew-mod}) is a positively-biased version of eq(\ref{eq:piecew}), which has its practical importance due to the fact that the last term in eq(\ref{eq:piecew}) from Theorem \ref{thm:piecew} results in zero-mean, noisy gradients, when the local functions are nearing their local solutions, when in fact, this is the area where more local work is not needed.  Thus, a positive bias is desirable to drive the number of local steps down.  This result is also useful from a communication efficiency perspective in its implementation, because the server has all the components to calculate this quantity, and would not require additional communication.

\subsubsection{Regularization for Number of Local Steps} \label{sec:reg}
One of the goals for FATHOM is savings in local computation.  To avoid excessive number of local steps, we further develop a regularization term for local computation against excessive $K$, which is a proxy for the hypergradient of the local client functions at the end of each round : $\partial f_{i}(x_{t}^{i,K}) / \partial K$.

\begin{theoremrep} \label{thm:piecew2}
When a piecewise function $J_{t}$ is defined for every value of $K_{0} \in [K]$ on $l$, such that $0.0 \leq l < 1.0$, we claim, under Assumption \ref{assump:convex}, that the following is a subgradient of $\sum_{i=1}^{m} f_{i}(x_{t}^{i,K_{t}})$ at $K_{t} = K_{0}$:
\begin{equation} \label{eq:piecew2}
    \frac{\partial J_{t}}{\partial l} = -\eta_{L,t}\sum_{i=1}^{m} \frac{\nu_{i}}{\nu} \mathbb{E}\big[g_{i}(x_{t}^{i, K_{0}-1})\big] \cdot g_{i}(x_{t}^{i, K_{t}}) \approx -\eta_{L,t} \sum_{i=1}^{m} \frac{\nu_{i}}{\nu} \sum_{k=0}^{K_{t}-1} g_{i}(x_{t}^{i, k}) \cdot g_{i}(x_{t}^{i, K_{t}}) 
\end{equation}
where $l$ represents the marginal fraction of local steps beyond $K_{0}$.  We leave the proof in the Appendix section beginning in eq(\ref{eq:piecew2-j}).
\end{theoremrep}
\begin{proof}
We clarify this quantity by re-writing $J_{K_{0}}(l) = J_{t}$ and its definition (with some abuse of notation):
\begin{equation} \label{eq:piecew2-j}
    J_{K_{0}}(l) = \sum_{i=1}^{m}\frac{\nu_{i}}{\nu} f_{i}(x_{t}^{i,K-1}) \bigg|_{K = K_{0} + l} =  \sum_{i=1}^{m}\frac{\nu_{i}}{\nu} f_{i} \Big( x_{t}^{i,K_{0}-1} - l\big(\eta_{L,t} g_{i}(x_{t}^{i,K_{0}}) \big) \Big)
\end{equation}
for $K_{0} \geq 1$, where $l$ represents the marginal fraction of local steps $K_{0}$.  Then, by convexity from Assumption \ref{assump:convex}, and by recalling that $l$ represents the marginal fraction of local steps, we claim that:
\begin{equation}
    \frac{\partial J_{K_{0}}(l)}{\partial l} \leq \Big[ \frac{\sum_{i=1}^{m}\frac{\nu_{i}}{\nu} \big( f_{i}(x_{t}^{i,K-1}) \big|_{K = K_{0} + l} - f_{i}(x_{t}^{i,K-1}) \big|_{K = K_{0} } \big)}{l} \Big]
\end{equation}
From this result, we conclude that $\frac{\partial J_{K_{0}}(l)}{\partial l}$ is a subgradient of $f(x_{t})$ at $K = K_{0}$.
With abuse of notation by setting $K_{0} = K_{t}$ and therefore $L_{t} = L_{K_{0}}(l)$, we further derive $\frac{\partial J_{t}}{\partial l}$ by breaking it down similarly as eq(\ref{eq:hyp1}), leads us to:
\begin{align}
    \frac{\partial J_{t}}{\partial l} & = \sum_{i=1}^{m} \frac{\nu_{i}}{\nu} \nabla f_{i}(x_{t}^{i, K_{t}-1}) \cdot \frac{\partial \big(- l\eta_{L,t} g_{i}(x_{t}^{i, K_{t}})\big)}{\partial l}   \\
    & \overset{\mathrm{(J1)}}{=} -\eta_{L,t}\sum_{i=1}^{m}\frac{\nu_{i}}{\nu}  \mathbb{E}\big[g_{i}(x_{t}^{i, K_{0}-1})\big] \cdot g_{i}(x_{t}^{i, K_{t}}) \\
    & \overset{\mathrm{(J2)}}{\approx} -\eta_{L,t} \sum_{i=1}^{m}\frac{\nu_{i}}{\nu} \sum_{k=0}^{K_{t}-1} g_{i}(x_{t}^{i, k}) \cdot g_{i}(x_{t}^{i, K_{t}}) 
\end{align}
where J1 follows from Assumption \ref{assump:unbiased}, and J2 crudely assumes $\nabla f_{i}(x_{t}^{i, K_{t}-1}) \approx \sum_{k=0}^{K_{t}-1} g_{i}(x_{t}^{i, k})$ from the additional averaging.  The above result concludes the proof for Theorem \ref{thm:piecew2}.
\end{proof}

In our algorithm, we use the normalized update based on the following biased proxy, since eq(\ref{eq:piecew2}) tends to be noisy from $g_{i}(x_{t}^{i, K_{t}})$.
\begin{align} 
    G_{t} & = -\eta_{L,t} \sum_{i=1}^{m} \frac{\nu_{i}}{\nu} \min_{K \leq K_{t}} \Big(  \sum_{k=0}^{K-1} g_{i}(x_{t}^{i, k}) \cdot g_{i}(x_{t}^{i, K}) \Big) \label{eq:hypreg2} \\
    \overline{G}_{t} & = -\eta_{L,t} \sum_{i=1}^{m} \frac{\nu_{i}}{\nu} \min_{K \leq K_{t}} \Bigg(  \frac{\sum_{k=0}^{K-1} g_{i}(x_{t}^{i, k})}{\big\|\sum_{k=0}^{K-1} g_{i}(x_{t}^{i, k})\big\|} \cdot \frac{g_{i}(x_{t}^{i, K})}{\|g_{i}(x_{t}^{i, K})\|} \Bigg) \label{eq:hypreg2norm}
\end{align}
where $\overline{G}_{t}$ is the normalized update.  The proxy yields a bias towards smaller number of local steps, which is desirable for reducing local computation.  We use this biased proxy against using a more typical regularization such as L2 for the number of local steps, based on initial empirical results for better performance..

\subsubsection{Normalized Exponentiated Gradient Updates} \label{sec:updates}
For the update rules of the hyperparameters $\eta_{L}$ (client learning rate) and $K$ (number of client local steps), we use the normalized exponentiated gradient descent method (EGN) with no momentum, rather than a conventional linear update method such as the additive update of hypergradient descent proposed in \citet{baydin2017online}.  It is reasonable to use exponentiated gradient (EG) methods for updates of hyperparameters that are strictly positive in value.  EG methods also enjoy significantly faster convergence properties when only a small subset of the dimensions are relevant, according to \citet{amid2022stepsize}.  

EG methods have been proposed in previous works for a variety of applications (\citet{khodak2021federated}, \citet{amid2022stepsize}, \citet{li2020geometryaware}), and analyzed in depth (\citet{ghai2019exponentiated}), where its convergence has been studied and validated (\citet{2018egmconv}).  Recently, \citet{amid2022stepsize} showed that EGN is the same as the multiplicative update for hypergradient descent proposed in \citet{baydin2017online}, when the approximation $exp(\boldsymbol{\cdot}) \approx 1+\boldsymbol{\cdot}$ is made.  From our observations, we believe that momentum is not needed for the effectiveness of EGN in our application, as validated in our numerical results.  We also opted-out of adding further complexity such as extra weights and activation functions to model the relationships between $\eta_{L,t}$ and $K_{t}$, because it would require more samples to optimize and because FATHOM is a one-shot procedure. Furthermore, due to the non-stationary nature of these values, we opt for a simpler scheme for faster performance.  

Hence, for the update rule of client learning rate, $\eta_{L}$, we have:
\begin{equation} \label{eq:etaupdate}
    \eta_{L, t+1} = \eta_{L, t} \exp\big(-\gamma_{\eta} \overline{H}_{t}\big)
\end{equation}
where $\overline{H}_{t}$ is as defined in eq(\ref{eq:hyp1approx}).  For number of local steps, we observe that it is related to batch size in round $t$, $B_{t}$, as follows.  To accommodate heterogeneity of local dataset sizes among clients, we have number of local data samples from client $i$ to be $\nu_{i}$.  The number of local steps for client $i$ is $K_{i} =  \lfloor \nu_{i} E_{t} / B_{t} \rfloor$, where $E_{t}$ is number of epochs, with $E_{t} = 1$ meaning the entire local dataset for each client to be processed once per round.  We derive update rules for $E_{t}$ and $B_{t}$ globally to optimize the number of local steps, without having to make any changes to our theoretical analysis to accommodate the heterogeneity of local dataset sizes:
\begin{equation} \label{eq:Eupdate}
    E_{t+1} = E_{t} \exp\big(-\gamma_{E} \big(\ \overline{N}_{t} + \overline{G}_{t}\big) \big)
\end{equation} 
and
\begin{equation} \label{eq:Bupdate}
    B_{t+1} = B_{t} \exp\big(-\gamma_{B} \big(- \overline{G}_{t}\big) \big)
\end{equation}
where $N_{t}$ and $G_{t}$ are defined in eq(\ref{eq:piecew-modapprox}) and eq(\ref{eq:hypreg2norm}), respectively.  These update rules accomplish the goal of updating the number of local steps via $E_{t} / B_{t}$ with $\frac{E_{t+1}}{B_{t+1}} = \frac{E_{t}}{B_{t}} \exp\big(-\gamma_{E} \overline{N}_{t} - \big(\gamma_{E} - \gamma_{B}\big) \overline{G}_{t}\big) $. Typically, with $\gamma_{B} \geq \gamma_{E}$, $(\gamma_{B} - \gamma_{E})\overline{G}_{t}$ becomes a tunable regularization term as discussed at the end of Section \ref{sec:reg}.

\subsubsection{Client Sampling}
We present our method, FATHOM, as shown in Algorithm \ref{alg:fathom2}. One practical factor we have not considered in our discussions is partial client sampling.  For our implementation to handle the stochastic nature of client sampling, the metric $\overline{\Delta}_{t-1}$ for calculating $\overline{H}_{t}$ in eq(\ref{eq:hyp1approx}) and $\overline{N}_{t}$ in eq(\ref{eq:piecew-modapprox}) is modified by a smoothing function for noise filtering, i.e. $\overline{\Delta}_{t, sm} = \alpha \overline{\Delta}_{t-1, sm} + (1 - \alpha) \overline{\Delta}_{t}$, which is a single-pole infinite impulse response filter (\citet{10.5555/1795494}Oppenheimer et al. [2009]) with no bias compensation. We use the notation "sm" for smoothed, and after many experiments, we decide to use $\alpha = 0.5$ for all of our numerical results.

\begin{algorithm}[H] \label{alg:fathom2}
\SetAlgoLined
\KwIn{Server initializes global model $x_{t=1}$, $T$ as the end communication round, and:
\begin{equation*}\overline{\Delta}_{t=0, sm} = 0\ \text{;} \ \alpha = 0.5 \ \text{;} \ \gamma_{\eta} = 0.01 \ \text{;} \ \gamma_{E} = 0.01 \ \text{;} \ \gamma_{B} = 0.1\end{equation*}
}
\KwOut{$x_{T}$, as well as $\eta_{L, t}$, $E_{t}$ and $B_{t}$ for all $t \in [T]$}

\For{$t = 1, \dotsc, T$}{
    Sample client set $S_{t}$ out of $m$ clients.

    For each client $i \in S_{t}$, initialize: $x_{t}^{i, k=0} = x_{t}$ and $K_{t,i} = \lfloor \nu_{i}E_{t} / B_{t \rfloor}$ . 
    
    Set $\Delta_{i} = 0$, and $\phi_{i} = +\infty$.
    
	\For {$k = 0, \dotsc, K_{t,i}-1$}{
        For each client $i$, compute in parallel an unbiased stochastic gradient $g_{i}(x_{t}^{i, k})$.
        
        For each client $i$, calculate $\phi_{i} = \min(\phi_{i}, g_{i}(x_{t}^{i, k}) \cdot \Delta_{i})$ where $\Delta_{i} = x_{t}^{i, k} - x_{t}$ 
        
        For each client $i$, update in parallel its local solution: $x_{t}^{i, k+1} = x_{t}^{i, k} - \eta_{L, t} g_{i}(x_{t}^{i, k})$
	}
	Server calcualtes $\nu = \sum_{i \in S_{t}} \nu_{i}$, where $\nu_{i}$ is the size of client $i$ dataset.
	
	Server calculates $\overline{\Delta}_{t} = \sum_{i \in S_{t}} \Delta_{i} (\nu_{i}/\nu)$; see eq(\ref{eq:deltabar})
	
	Server updates global model $x_{t+1} = x_{t} - \overline{\Delta}_{t}$

	Server calculates $\overline{H}_{t} = \overline{N}_{t} = -\frac{\overline{\Delta}_{t}}{\|\overline{\Delta}_{t}\|} \cdot \frac{\overline{\Delta}_{t-1,sm}}{\|\overline{\Delta}_{t-1,sm}\|}$, modified from eq(\ref{eq:hyp1approx}) and eq(\ref{eq:piecew-modapprox})

	Server calculates $\overline{G}_{t}$; see eq(\ref{eq:hypreg2norm}
	
	Server updates client learning rate $\eta_{L, t+1}$, epochs, $E_{t+1}$, and batch size $B_{t+1}$ for the next round; see eq(\ref{eq:etaupdate}), eq(\ref{eq:Eupdate}), and eq(\ref{eq:Bupdate}).
	
	Server updates $\overline{\Delta}_{t,sm} = (1 - \alpha) \overline{\Delta}_{t} + \alpha \overline{\Delta}_{t-1,sm}$ for the next round
}
\caption{\textbf{FATHOM} : $g_{i}(x)$ is defined in Assumptions \ref{assump:unbiased}, and $m$ is the number of clients. }
\end{algorithm}

\section{Theoretical Convergence} \label{sec:theory}
A standard approach to theoretical analysis of an online optimization method such as ours, is through analyzing the regret bound (\citet{10.5555/3041838.3041955}, \citet{khodak2019adaptive},  \cite{kingma2014adam}, and \citet{Mokhtari2016}).  Nonetheless, this approach does not tell us the impact on communication efficiency by the online updates introduced from FATHOM.  Therefore, we take an alternative approach by extending the guarantees of FedAvg performance (\citet{wang2021field},  \citet{reddi2020adaptive}, \citet{gorbunov2020local}, \citet{yang2021achieving}, \citet{li2019convergence}, etc) to include both adaptive learning rate and adaptive number of local steps.  We assume the special case in our analysis to have full client participation.  We prove that adaptive learning rate and adaptive number of local steps does not impact asymptotic convergence, despite the given relaxed conditions.  

\begin{toappendix}
This supplementary section contains all the missing proofs from Section \ref{sec:theory}.
\end{toappendix}

\subsection{Assumptions} \label{sec:assump}
\begin{assumption} \label{assump:lip}
(L-Lipschitz Continuous Gradient for Parameters $x_t$) There exists a constant $L > 0$, such that $\| \nabla f_{i}(x) - \nabla f_{i}(y) \| \leq L \| x-y \|$, $\forall x$, $y \in \mathbb{R}^{d}$, and $i \in [m]$, where $x$ and $y$ are the parameters in eq(\ref{eq:erm}.
\end{assumption}
\begin{assumption} \label{assump:boundedvar}
(Bounded Local Variance) There exist a constant $\sigma_{L} > 0$, such that the variance of each local gradient estimator is bounded by $\mathbb{E} \| \nabla f_{i}(x) - g_{i}(x) \|^{2} \leq \sigma_{L}^{2}$, $\forall x$, and $i \in [m]$.  
\end{assumption}
\begin{assumption} \label{assump:boundedsec}
(Bounded Second Moment) There exists a constant $G > 0$, such that $\mathbb{E}_{t}\|\nabla f_{i}(x_{t})\| \leq G$, $i \in [m]$, $\forall x_{t}$.
\end{assumption}

\subsection{Convergence Results}

\begin{theoremrep}  \label{thm:fathom-bound}
Under Assumptions \ref{assump:unbiased}-\ref{assump:boundedsec} and with full client participation, when FATHOM as shown in Algorithm \ref{alg:fathom2} is used to find a solution $x_{*}$ to the unconstrained problem defined in eq(\ref{eq:erm}), the sequence of outputs $\{x_{t}\}$ satisfies the following upper-bound, where, with slight abuse of notation, $\mathcal{E} = \min_{t \in [T]} \mathbb{E}_{t} \|\nabla f(x_{t}) \|_{2}^{2} $:
\begin{equation} \label{eq:epsobound-fathom}
    \mathcal{E}_{fathom} = \mathcal{O}\bigg( \sqrt{\frac{\sigma_{L}^{2}+G^{2}}{m\overline{K}T}}  + \sqrt[3]{\frac{\sigma_{L}^{2}}{\overline{K}T^{2}}} + \sqrt[3]{\frac{G^{2}}{T^{2}}}\bigg)
\end{equation} with the following conditions: $\overline{\eta}_{L} = \min \Bigg( \sqrt{\frac{2\beta_{0}mD}{\beta_{1}\overline{K}LT(\sigma_{L}^{2} + G^{2})}}, \sqrt[3]{\frac{\beta_{0}D}{2.5\beta_{2}\overline{K}^{2}L^{2}\sigma_{L}^{2}T}}, \sqrt[3]{\frac{\beta_{0}D}{2.5\beta_{3}\overline{K}^{3}L^{2}G^{2}T}} \Bigg)$ and $\eta_{L,t} \leq 1/L$ for all $t$, where 
\begin{equation} \label{eq:adaptivecond}
    \overline{\eta}_{L} \triangleq \frac{1}{T} \sum_{t=1}^{T} \eta_{L,t} \qquad \text{and} \qquad \overline{K} \triangleq \frac{1}{T} \sum_{t=1}^{T} K_{t}
\end{equation} and where
\begin{align}
    \beta_{0} = \frac{\sum_{t}\eta_{L,t} \label{eq:beta0beta1} K_{t}}{T[\frac{1}{T}\sum_{t}\eta_{L,t}][\frac{1}{T}\sum_{t}K_{t}]} & \  \text{ , } \  \beta_{1} = \frac{\sum_{t}\eta_{L,t} K_{t}\big[ \frac{1}{T}\sum_{t}\eta_{L,t}\big]}{\sum_{t}\eta_{L,t}^{2} K_{t}} \\
    \beta_{2} = \frac{\sum_{t}\eta_{L,t} \label{eq:beta2beta3} K_{t} \big[ \frac{1}{T}\sum_{t}\eta_{L,t}\big]^{2}\big[ \frac{1}{T}\sum_{t}K_{t}\big]}{\sum_{t}\eta_{L,t}^{3} K_{t}^{2}} & \   \text{ , } \  \beta_{3} = \frac{\sum_{t}\eta_{L,t} K_{t} \big[ \frac{1}{T}\sum_{t}\eta_{L,t}\big]^{2}\big[ \frac{1}{T}\sum_{t}K_{t}\big]^{2}}{\sum_{t}\eta_{L,t}^{3} K_{t}^{3}}
\end{align}
We leave the proof in the Appendix beginning in eq(\ref{eq:thm1proof4}).
\end{theoremrep}
\begin{proof} 
We begin by first defining the following:

By re-writing eq(\ref{eq:thm1proof3}) from Lemma \ref{lem:fixedhyp} with adaptive $\eta_{L, t}$ and $K_{t}$, we end up with:
\begin{equation}
    \sum_{t=0}^{T-1}\frac{\eta_{L,t}K_{t}}{2}\mathbb{E}_{t}\big\|\nabla f(x_{t})\big\|^{2} \leq f(x_{0}) - f(x_{T}) + \sum_{t=0}^{T-1} \eta_{L,t}K_{t} \big[ \frac{\eta_{L,t}L}{2m} \big( \sigma_{L}^{2} + G^{2} \big) + \frac{5\eta_{L,t}^{2}K_{t}L^{2}}{2}\big(\sigma_{L}^{2} + K_{t}G^{2}\big) \big] \label{eq:thm1proof4} 
\end{equation}
After re-arranging:
\begin{equation}
    \min_{t\in[T]} \mathbb{E}_{t}\|\nabla f(x_{t}) \|^{2} \leq \underbrace{\frac{2D}{\sum_{t}\eta_{L,t}K_{t}}}_\text{progress} + \underbrace{\frac{L\sum_{t}\eta_{L,t}^{2}K_{t}}{m\sum_{t}\eta_{L,t}K_{t}}\big( \sigma_{L}^{2} + G^{2} \big)}_\text{deviation 1} + \underbrace{\frac{5L^{2}\sum_{t}\eta_{L,t}^{3}K_{t}^{2}}{\sum_{t}\eta_{L,t}K_{t}}\sigma_{L}^{2}}_\text{deviation 2} + \underbrace{\frac{5L^{2}\sum_{t}\eta_{L,t}^{3}K_{t}^{3}}{\sum_{t}\eta_{L,t}K_{t}}G^{2}}_\text{deviation 3} \label{eq:thm1proof5}    
\end{equation}
which is followed by:
\begin{equation} \label{eq:epsbound-fathom}
    \mathcal{E}_{fathom} \leq 
    \underbrace{\frac{2\beta_{0}D}{\overline{\eta}_{L}\overline{K}T}}_\text{progress} + \underbrace{\frac{\beta_{1}\overline{\eta}_{L}L}{m}\big( \sigma_{L}^{2} + G^{2} \big)}_\text{deviation 1} + \underbrace{5\beta_{2}\overline{\eta}_{L}^{2}\overline{K}L^{2}\sigma_{L}^{2}}_\text{deviation 2} + \underbrace{5\beta_{3}\overline{\eta}_{L}^{2}\overline{K}^{2}L^{2}G^{2}}_\text{deviation3}
\end{equation}
We have one progress term, and three deviation terms, similar to the labeling scheme in the convex result from \citet{wang2021field}Wang et al. [2021].  Typically, one of these terms dominates during the course of the optimization process, where it is desirable to never let one of the deviation terms to become dominant.  When we set each of the deviation terms to be equal to the progress term, we recover the bound shown in eq(\ref{eq:epsobound-fathom} when the conditions are met.  This concludes the proof for Theorem \ref{thm:fathom-bound}.  
\end{proof}
\begin{toappendix}
\begin{lemmarep} \label{lem:fixedhyp}
Under Assumptions \ref{assump:unbiased}-\ref{assump:boundedsec} and with full client participation, when FedAvg with constant hyperparameters is used to find a solution $x_{*}$ to the unconstrained problem defined in eq(\ref{eq:erm}), the sequence of outputs $\{x_{t}\}$ satisfies the following upper-bound, where, with slight abuse of notation, $\mathcal{E} = \min_{t \in [T]} \mathbb{E}_{t} \|\nabla f(x_{t}) \|_{2}^{2} $:
\begin{equation}
    \min_{t\in[T]} \mathbb{E}_{t}\|\nabla f(x_{t}) \|^{2} \leq \underbrace{\frac{2D}{\eta_{L}KT}}_\text{progress} + \underbrace{\frac{\eta_{L}L}{m}\big( \sigma_{L}^{2} + G^{2} \big)}_\text{deviation 1} + \underbrace{5\eta_{L}^{2}KL^{2}\sigma_{L}^{2}}_\text{deviation 2} + \underbrace{5\eta_{L}^{2}K^{2}L^{2}G^{2}}_\text{deviation3} \label{eq:thm1proof2}
\end{equation} where $D = f(x_{0}) - f(x_{T}) = f(x_{0}) - f(x_{*})$ with $x_{*}$ being the fixed point solution discussed in Section \ref{sec:assump}.  Eq(\ref{eq:thm1proof2}) has one progress term and three deviation terms, where $\eta_{L} \leq \frac{1}{L}$ needs to hold for client local gradient descent to guarantee local progress.  
\end{lemmarep}
\begin{proof}
We start proving convergence of the non-convex problem by bounding the progress made in the loss function within a single round, loosely following the beginning steps from the Proof of Theorem 1 in \citet{yang2021achieving} Yang et al [2021]:
\begin{align}
    \mathbb{E}_{t}[f(x_{t+1}] & \leq \mathbb{E}_{t}[f(x_{t}] + \langle\nabla f(x_{t}), \mathbb{E}_{t}(x_{t+1} - x_{t})\rangle + \frac{L}{2}\mathbb{E}_{t} \| x_{t+1} - x_{t} \|^{2} \\
    & = \mathbb{E}_{t}[f(x_{t}] + \langle\nabla f(x_{t}), \mathbb{E}_{t}[\overline{\Delta}_{t} + \eta_{L}K\nabla f(x_{t}) - \eta_{L}K\nabla f(x_{t})])\rangle + \frac{L}{2}\mathbb{E}_{t} \| \overline{\Delta}_{t} \|^{2} \\
    & = \mathbb{E}_{t}[f(x_{t}] - \eta_{L}K\|\nabla f(x_{t})\|^{2} + \underbrace{\langle\nabla f(x_{t}), \mathbb{E}_{t}[\overline{\Delta}_{t} + \eta_{L}K\nabla f(x_{t})] \rangle }_{A_{1}} + \frac{L}{2}\underbrace{\mathbb{E}_{t} \| \overline{\Delta}_{t} \|^{2}}_{A_{2}}
\end{align}
By using results from Lemma \ref{lem:a1} and Lemma \ref{lem:a2}, we have what follows:
\begin{equation} \label{eq:thm1proof1}
    \mathbb{E}_{t}[f(x_{t+1}] \leq \mathbb{E}_{t}[f(x_{t}] - \frac{\eta_{L}K}{2} \| \nabla f(x_{t}) \|^{2} + \frac{5\eta_{L}^{3}K^{2}L^{2}}{2} \Big( \sigma_{L}^{2} + KG^{2} \Big) + \frac{L\eta_{L}^{2}K}{2m} \Big( \sigma_{L}^{2} + G^{2} \Big)
\end{equation}
Therefore, in order to guarantee progress in each round, the following condition is required to hold true:
\begin{equation}
    \| \nabla f(x_{t}) \|^{2} \geq 5\eta_{L}^{2}KL^{2} \Big( \sigma_{L}^{2} + KG^{2} \Big) + \frac{L\eta_{L}}{m} \Big( \sigma_{L}^{2} + G^{2} \Big)
\end{equation}
Continuing from eq(\ref{eq:thm1proof1}), and summing telescopically, we end up with:
\begin{equation}
    \sum_{t=0}^{T-1}\frac{\eta_{L}K}{2}\mathbb{E}_{t}\big\|\nabla f(x_{t})\big\|^{2} \leq f(x_{0}) - f(x_{T}) + T \eta_{L}K \big[ \frac{\eta_{L}L}{2m} \big( \sigma_{L}^{2} + G^{2} \big) + \frac{5\eta_{L}^{2}KL^{2}}{2}\big(\sigma_{L}^{2} + KG^{2}\big) \big] \label{eq:thm1proof3} 
\end{equation} where $D = f(x_{0}) - f(x_{T}) = f(x_{0}) - f(x_{*})$ with $x_{*}$ being the fixed point solution discussed in Section \ref{sec:assump}.  This concludes the proof of Lemma \ref{lem:fixedhyp}.
\end{proof}

\begin{lemmarep} \label{lem:a1}
Under Assumptions \ref{assump:unbiased}-\ref{assump:boundedsec} and with full client participation, we claim the following is true:
\begin{equation}
    A_{1} \leq \frac{\eta_{L}K}{2} \|\nabla f(x) \|^{2} + \frac{5K^{2}\eta_{L}^{3}L^{2}}{2}\big(\sigma_{L}^{2} + KG^{2} \big)
\end{equation}
\end{lemmarep}
\begin{proof}
We start by following most of the initial steps from the Proof of Theorem 1 in \citet{yang2021achieving} Yang et al [2021]:
\begin{align}
A_{1} & = \langle\nabla f(x_{t}), \mathbb{E}_{t}[\overline{\Delta}_{t} + \eta_{L}K\nabla f(x_{t})] \rangle \\
& = \Big\langle\nabla f(x_{t}), \mathbb{E}_{t}\Big[-\frac{1}{m}\sum_{i=1}^{m}\sum_{k=0}^{K-1}\eta_{L}g_{i}(x_{t}^{i,k}) + \eta_{L}K\nabla f(x_{t})\Big] \Big\rangle \\
& = \Big\langle\nabla f(x_{t}), \mathbb{E}_{t}\Big[-\frac{1}{m}\sum_{i=1}^{m}\sum_{k=0}^{K-1}\eta_{L}\nabla f_{i}(x_{t}^{i,k}) + \eta_{L}K\frac{1}{m}\sum_{i=1}^{m}\nabla f_{i}(x_{t})\Big] \Big\rangle \\
& = \Big\langle \sqrt{\eta_{L}K} \nabla f(x_{t}), -\frac{\sqrt{\eta_{L}}}{m\sqrt{K}}\mathbb{E}_{t}\sum_{i=1}^{m}\sum_{k=0}^{K-1}(\nabla f_{i}(x_{t}^{i,k}) - \nabla f_{i}(x_{t})) \Big\rangle \\
& \overset{\mathrm{(a1)}}{=} \frac{\eta_{L}K}{2}\|\nabla f(x_{t})\|^{2} + \frac{\eta_{L}}{2Km^{2}}\mathbb{E}_{t}\Big\|\sum_{i=1}^{m}\sum_{k=0}^{K-1}\nabla f_{i}(x_{t}^{i,k}) - \nabla f_{i}(x_{t})\Big\|^{2} - \frac{\eta_{L}}{2Km^{2}}\mathbb{E}_{t}\Big\|\sum_{i=1}^{m}\sum_{k=0}^{K-1}\nabla f_{i}(x_{t}^{i,k})\Big\|^{2} \\
& \overset{\mathrm{(a2)}}{\leq} \frac{\eta_{L}K}{2}\|\nabla f(x_{t})\|^{2} + \frac{\eta_{L}}{2m}\sum_{i=1}^{m}\sum_{k=0}^{K-1}\mathbb{E}_{t}\big\|\nabla f_{i}(x_{t}^{i,k}) - \nabla f_{i}(x_{t})\big\|^{2} - \frac{\eta_{L}}{2Km^{2}} \sum_{i=1}^{m}\sum_{k=0}^{K-1}\mathbb{E}_{t}\Big\|\nabla f_{i}(x_{t}^{i,k})\Big\|^{2} \\
& \overset{\mathrm{(a3)}}{\leq} \frac{\eta_{L}K}{2}\|\nabla f(x_{t})\|^{2} + \frac{\eta_{L}L^{2}}{2m}\sum_{i=1}^{m}\sum_{k=0}^{K-1}\mathbb{E}_{t}\big\|x_{t}^{i,k} - x_{t}\big\|^{2} - \frac{\eta_{L}}{2m}G^{2} \label{eq:lem1-last}\\
& \overset{\mathrm{(a4)}}{\leq} \frac{\eta_{L}K}{2}\|\nabla f(x_{t})\|^{2} + \frac{5K^{2}\eta_{L}^{3}L^{2}}{2}\big(\sigma_{L}^{2} + KG^{2}\big)
\end{align} where, from \citet{yang2021achieving} Yang et al [2021], (a1) follows from that  $\langle x,y \rangle = \frac{1}{2}[\|x\|^{2} + \|y\|^{2} - \|x-y\|^{2}]$ for $x = \sqrt{\eta_{L}K}\nabla f(x_{t})$ and $y = -\frac{\sqrt{\eta_{L}}}{m\sqrt{K}}\sum_{i=1}^{m}\sum_{k=0}^{K-1}(\nabla f_{i}(x_{t}^{i,k}) - \nabla f_{i}(x_{t}))$, (a2) is due to $\mathbb{E}\|x_{1} + x_{2} + \dots + x_{n}\|^{2} \leq n\mathbb{E}[\|x_{1}\|^{2}+\|x_{2}\|^{2}+ \dots +\|x_{n}\|^{2}]$ and $\mathbb{E}[\|x_{1}\|^{2}+\|x_{2}\|^{2}+ \dots +\|x_{n}\|^{2}] \leq \mathbb{E}\|x_{1} + x_{2} + \dots + x_{n}\|^{2}$, (a3) is due to Assumption \ref{assump:lip}, which is where we start to diverge from \citet{yang2021achieving} Yang et al [2021].  Our result from Lemma \ref{lem:a1a} by using Assumption \ref{assump:boundedsec}, combined with removal of the last term, justifies (a4) above, and thus concludes the proof for Lemma \ref{lem:a1}.  The last term of eq(\ref{eq:lem1-last}) could have remained for a tighter final bound in the theorems, but would require to restrict $K$ such that $\eta_{L}K \leq \frac{1}{L}$ which we try to avoid.

\end{proof}
\begin{lemmarep} \label{lem:a2}
Under Assumptions \ref{assump:unbiased}-\ref{assump:boundedsec} and with full client participation, we claim the following is true:
\begin{equation}
    A_{2} \leq \frac{\eta_{L}K^{2}}{m}\big[\sigma_{L}^{2} + G^{2}\big]
\end{equation}
\end{lemmarep}
\begin{proof}
We start with the following definition of $\overline{\Delta}_{t} = \frac{1}{m} \sum_{i=1}^{m} \Delta_{t}^{i} = \big( -\frac{\eta_{L}}{m} \sum_{i=1}^{m} \sum_{k=0}^{K_{t}-1} g_{i}(x_{t}^{i,k}) \big)$:
\begin{align}
  \mathbb{E}_{t} \big\| \overline{\Delta}_{t} \big\|^{2} & = \mathbb{E}_{t} \Big\| \frac{1}{m} \sum_{i=1}^{m} \Delta_{t}^{i} \Big\|^{2} \\
  & = \frac{1}{m^{2}} \mathbb{E}_{t} \Big\| \sum_{i=1}^{m} \Delta_{t}^{i} \Big\|^{2} =  \frac{\eta_{L}^{2}}{m^{2}} \mathbb{E}_{t} \Big\| \sum_{i=1}^{m} \sum_{k=0}^{K-1} g_{i}(x_{t}^{i,k}) \Big\|^{2} \\
  & = \frac{\eta_{L}^{2}}{m^{2}} \mathbb{E}_{t} \Big\| \sum_{i=1}^{m} \sum_{k=0}^{K-1} (g_{i}(x_{t}^{i,k}) - \nabla f(x_{t}^{i,k})) \Big\|^{2} + \frac{\eta_{L}^{2}}{m^{2}} \mathbb{E}_{t} \Big\| \sum_{i=1}^{m} \sum_{k=0}^{K-1} \nabla f(x_{t}^{i,k}) \Big\|^{2} \\
  & \leq \frac{\eta_{L}K^{2}}{m}\big[\sigma_{L}^{2} + G^{2}\big]
\end{align} which completes the proof of Lemma \ref{lem:a2}.
\end{proof}
\begin{lemmarep} \label{lem:a1a}
Under Assumptions \ref{assump:lip}-\ref{assump:boundedsec} and with full client participation, we claim the following is true:
\begin{equation}
    \frac{1}{m}\sum_{i=1}^{m} \mathbb{E}_{t}\big\|x_{t}^{i,k}-x_{t}\big\|^{2} \leq 5K \big[ K \eta_{L}^{2} G^{2} + \eta_{L}^{2} \sigma_{L}^{2} \big]
\end{equation}
\end{lemmarep}
\begin{proof}
We start by loosely following Lemma 3 from \citet{reddi2020adaptive} Reddi et al [2020]: 
\begin{align}
    \mathbb{E}_{t}\|x_{t}^{i,k}-x_{t}\|^{2} & = \mathbb{E}_{t}\|x_{t}^{i,k-1}-x_{t}-\eta_{L}g_{i}(x_{t}^{i,k-1})\|^{2} \\
    & = \mathbb{E}_{t}\|x_{t}^{i,k-1}-x_{t}-\eta_{L}(g_{i}(x_{t}^{i,k-1}) - \nabla f_{i}(x_{t}^{i,k-1}) + \nabla f_{i}(x_{t}^{i,k-1}))\|^{2} \\
    & \leq \big(1 + \frac{1}{K-1}\big) \mathbb{E}_{t} \big\|x_{t}^{i,k-1} - x_{t})\big\|^{2} + K \eta_{L}^{2} \mathbb{E}_{t}\big\| \nabla f_{i}(x_{t}^{i,k-1} \big\|^{2} + \eta_{L}^{2} (g_{i}(x_{t}^{i,k-1}) - \nabla f_{i}(x_{t}^{i,k-1}) \\
    & \leq \big(1 + \frac{1}{K-1}\big) \mathbb{E}_{t} \big\|x_{t}^{i,k-1} - x_{t})\big\|^{2} + K \eta_{L}^{2} G^{2} + \eta_{L}^{2} \sigma_{L}^{2}
\end{align}
The last two inequalities follows Assumption \ref{assump:boundedvar} and Assumption \ref{assump:boundedsec}, which yields a looser bound and which diverges from Lemma 3 from \citet{reddi2020adaptive} Reddi et al [2020]. Unrolling the recursion over $k$ and summing over clients $i \in [m]$:
\begin{align}
    \frac{1}{m}\sum_{i=1}^{m} \mathbb{E}_{t}\big\|x_{t}^{i,k}-x_{t}\big\|^{2} & \leq \sum_{p=0}^{K-1} \big(1 + \frac{1}{K-1}\big)^{p} \big[ K \eta_{L}^{2} G^{2} + \eta_{L}^{2} \sigma_{L}^{2} \big] \\
    & \leq K \big(1 + \frac{1}{K-1}\big)^{K} \big[ K \eta_{L}^{2} G^{2} + \eta_{L}^{2} \sigma_{L}^{2} \big] \\
    & \leq 5K \big[ K \eta_{L}^{2} G^{2} + \eta_{L}^{2} \sigma_{L}^{2} \big]
\end{align} where $\big(1 + \frac{1}{K-1}\big)^{K} \leq 5$ for $K > 1$. This concludes the proof of Lemma \ref{lem:a1a}.
\end{proof}

\end{toappendix}

The values of $\beta_{0}$, $\beta_{1}$, $\beta_{2}$, $\beta_{3}$, and $\beta_{4}$ are dependent on the relative changes over the adaptive process of these components, according to Chebyshev's Sum Inequalities (\citet{hardy1988inequalities}).  A special case is when these quantities equal to $1$ when both $\eta_{L,t}$ and $K_{t}$ are constant, which recovers the standard upperbound for FedAvg from eq(\ref{eq:epsobound-fathom}).

\begin{remark}\label{rem:relax}
The definitions in eq(\ref{eq:adaptivecond}) combined with the conditions for $\overline{\eta}_{L}$ above is called the relaxed conditions in this paper for the hyperparameters $\eta_{L, t}$ and $K_{t}$.
The values of $\eta_{L, t}$ and $K_{t}$ are adaptive during the optimization process between rounds $t=1$ and $t=T$, as long as the above conditions are satisfied for the guarantee in eq(\ref{eq:epsbound-fathom}) to hold.  This relaxation presents opportunities for a scheme such as FATHOM to exploit for performance gain. For example, suppose $T$ approaches $\infty$ for a prolonged training session. Then $\overline{\eta}_{L}$ would necessarily be sufficiently small for $\mathcal{E}_{fathom}$ to be bounded by eq(\ref{eq:epsobound-fathom}).  However, for early rounds i.e. small $t$ values, $\eta_{L,t} \leq T \overline{\eta}_{L}$ can be reasonably large and still can satisfy eq(\ref{eq:adaptivecond}), for the benefit of accelerated learning and convergence progress early on.  Similar strategy can be used for number of local steps to minimize local computations towards later rounds.  In any case, these strategies are mere guidelines meant to remain within the worst case guarantee.  However, Theorem \ref{thm:fathom-bound} offers the flexibility otherwise not available.  We will now show the empirical performance gained by taking advantage of this flexibility. 
\end{remark}

\begin{figure} [H]
  \centering
  \makebox[0pt]{\includegraphics[scale=0.29]{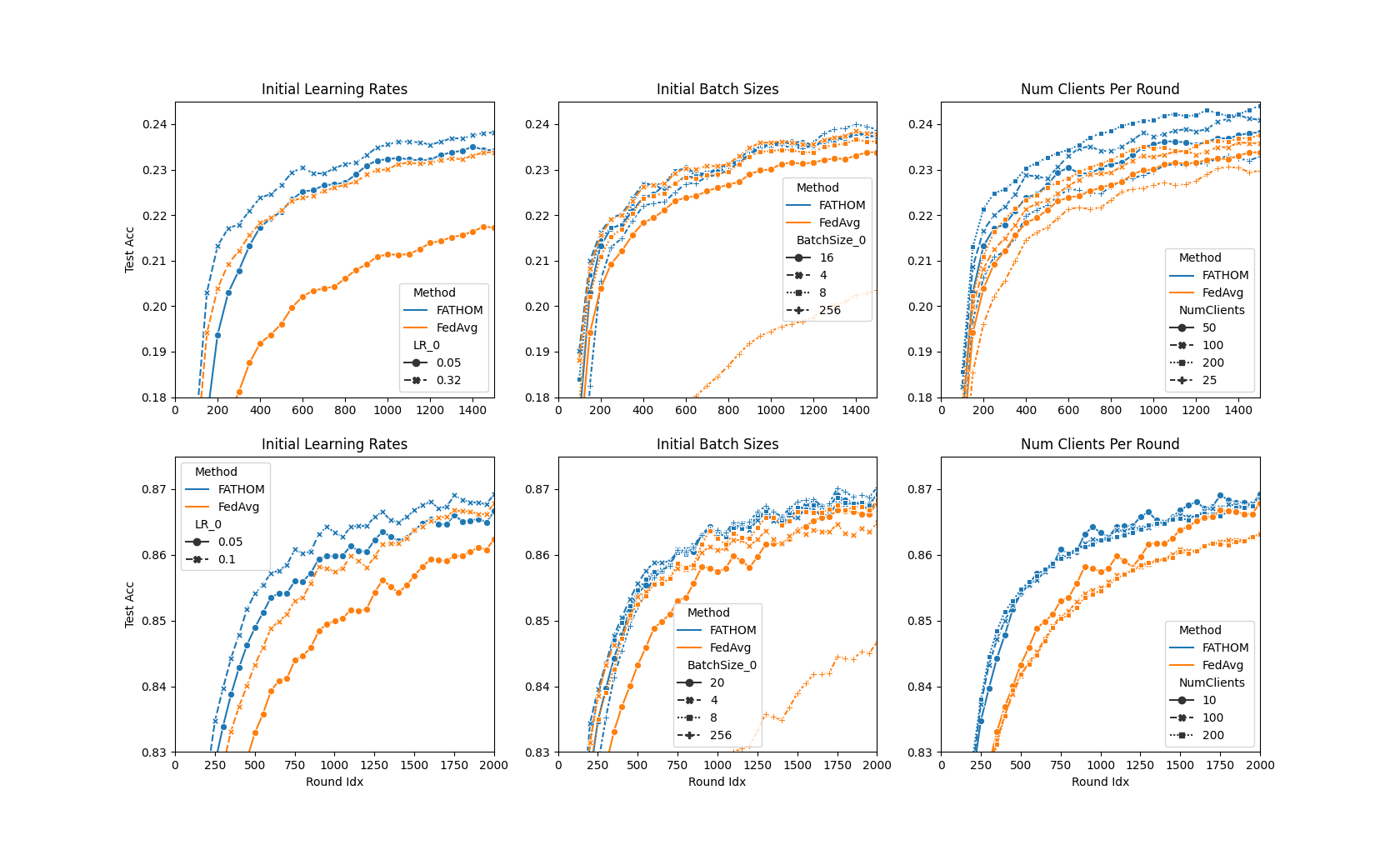}} 
  \caption{Test Accuracy Performance with various values of initial client learning rate (LR\_0), initial batch size (BatchSize\_0), and number of clients per round (NumClients). Top row: FSO sims. Bottom row: FEMNIST sims.  Baseline values for FEMNIST: LR\_0=0.1, BatchSize\_0=20, NumClients=10.  Baseline values for FSO: LR\_0=0.32, BatchSize\_0=16, NumClients=50. } \label{fig:fig1}
\end{figure}

\section{Empirical Evaluation and Numerical Results} \label{sec:sims}

We present an empirical evaluation of FATHOM proposed in Section \ref{sec:methodology} and outlined in Algorithm \ref{alg:fathom2}.  We conduct extensive simulations of federated learning in character recognition on the federated EMNIST-62 dataset (FEMNIST) (\citet{cohen2017emnist}) with a CNN, and in natural language next-word prediction on the federated Stack Overflow dataset (FSO) (\citet{authors2019tensorflow_so}) with a RNN.  We defer most of the details of the experiment setup in Appendix Section \ref{sec:app-sims}.  Our choice of datasets, tasks and models, are exactly the same as the "EMNIST CR" task and the "SO NWP" task from \citet{reddi2020adaptive}.  See Figure \ref{fig:fig1} and Table \ref{tbl:table1} and their captions for details of the experiment results.  Our evaluation lacks comparison with a few one-shot FL HPO methods discussed earlier in the paper because of a lack of standardized benchmark (until FedHPO-
B \citet{wang2022fedhpob} was published concurrently as this work) to be fair and comprehensive.

The underlying principle behind these experiments is evaluating the robustness of FATHOM versus FedAvg under various initial settings, to mirror realistic usage scenarios where the optimal hyperparameter values are unknown.  For FATHOM, we start with the same initial hyperparameter values as FedAvg.  The test accuracy progress with respect to communication rounds is shown in Figure \ref{fig:fig1} from these experiments.  We also pick test accuracy targets for the two tasks.  For FEMNIST CR we use 86\% and for FSO NWP we use 23\%.  Table \ref{tbl:table1} shows a table of resource utilization metrics with respect to reaching these targets in our experiments, highlighting the communication efficiency as well as reduction in local computation from FATHOM in comparison to FedAvg.  To our knowledge, we are the first to show gain from an online HPO procedure over a well-tuned equivalent procedure with fixed hyperparameter values.

The federated learning simulation framework on which we build our algorithms for our experiments is FedJAX (\citet{ro2021fedjax}) which is under the Apache License.  The server that runs the experiments is equipped with Nvidia Tesla V100 SXM2 GPUs.

\begin{table}[H]
  \caption{ Resource utilization in communication and local computation to reach specified test accuracy target for each task.  All evalutions are run for ten trials.  Bold numbers highlight better performance.  NA means target was not reached within 1500 rounds for FSO NWP and 2000 rounds for FEMNIST CR, in any of our trials.  LR\_0 is initial client learning rate, BS\_0 is initial batch size, and NCPR is number of clients per round.  All experiments use baseline initial values except where indicated.  For clarification, M is used in place for "million", and K for "thousand".\\Baseline\_fso : (LR\_0 = 0.32, BS\_0 = 16, NCPR = 50) \\Baseline\_femnist : (LR\_0 = 0.10, BS\_0 = 20, NCPR = 10)} \label{tbl:table1}
 
\begingroup
\setlength{\tabcolsep}{10pt} 
\renewcommand{\arraystretch}{1} 
\begin{adjustbox}{width=1.4\textwidth}
\centering

\begin{tabular}{llcccclllllllll}
\cline{1-6}
\multirow{2}{*}{Tasks}                                                            & \multirow{2}{*}{Experiments} & \multicolumn{2}{c}{\begin{tabular}[c]{@{}c@{}}Number of Rounds To\\ Reach Target Test Accuracy\end{tabular}} & \multicolumn{2}{c}{\begin{tabular}[c]{@{}c@{}}Local Gradients Calculated To\\ Reach Target Test Accuracy\end{tabular}} &  &  &  &  &  &  &  &  &  \\ \cline{3-6}
                                                                                  &                              & FATHOM                                                & FedAvg                                               & FATHOM                                                     & FedAvg                                                    &  &  &  &  &  &  &  &  &  \\ \cline{1-6}
\multirow{6}{*}{\begin{tabular}[c]{@{}l@{}}FSO NWP\\ Target@23\%\end{tabular}}    & Baseline\_fso                     & \textbf{562 $\pm$ 12}                                    & 971 $\pm$ 11                                            & \textbf{85M $\pm$ 1.2M}                                       & 124M $\pm$ 1.3M                                              &  &  &  &  &  &  &  &  &  \\
                                                                                  & LR\_0 = 0.05                 & \textbf{871 $\pm$ 7}                                     & NA                                                   & \textbf{138M $\pm$ 3.2M}                                               & NA                                                        &  &  &  &  &  &  &  &  &  \\
                                                                                  & BS\_0 = 4                    & 758 $\pm$ 43                                             & \textbf{580 $\pm$ 18}                                   & 93M $\pm$ 2.8M                                                & \textbf{74M $\pm$ 2.5M}                                      &  &  &  &  &  &  &  &  &  \\
                                                                                  & BS\_0 = 256                  & \textbf{801 $\pm$ 28}                                    & NA                                                   & \textbf{174M $\pm$ 18M}                                       & NA                                                        &  &  &  &  &  &  &  &  &  \\
                                                                                  & NCPR = 25                    & \textbf{970 $\pm$ 49}                                    & 1283 $\pm$ 33                                           & \textbf{63M $\pm$ 2.7M}                                       & 82M $\pm$ 3.8M                                               &  &  &  &  &  &  &  &  &  \\
                                                                                  & NCPR = 200                   & \textbf{396 $\pm$ 17}                                    & 684 $\pm$ 26                                            & \textbf{280M $\pm$ 45M}                                       & 350M $\pm$ 13M                                               &  &  &  &  &  &  &  &  &  \\ \cline{1-6}
\multirow{6}{*}{\begin{tabular}[c]{@{}l@{}}FEMNIST CR\\ Target@86\%\end{tabular}} & Baseline\_femnist                     & \textbf{739 $\pm$ 24}                                    & 1098 $\pm$ 15                                           & \textbf{1.5M $\pm$ 36K}                                       & 2.2M $\pm$ 64K                                               &  &  &  &  &  &  &  &  &  \\
                                                                                  & LR\_0 = 0.05                 & \textbf{905 $\pm$ 21}                                    & 1574 $\pm$19                                            & \textbf{1.7M $\pm$ 28K}                                       & 3.1M $\pm$ 28K                                               &  &  &  &  &  &  &  &  &  \\
                                                                                  & BS\_0 = 4                    & \textbf{708 $\pm$ 17}                                    & 885 $\pm$ 41                                            & \textbf{1.2M $\pm$ 28K}                                       & 1.7M $\pm$ 88K                                               &  &  &  &  &  &  &  &  &  \\
                                                                                  & BS\_0 = 256                  & \textbf{736 $\pm$ 20}                                    & NA                                                   & \textbf{2.0M $\pm$ 44K}                                       & NA                                                        &  &  &  &  &  &  &  &  &  \\
                                                                                  & NCPR = 100                   & \textbf{777 $\pm$ 16}                                    & 1436 $\pm$ 18                                           & \textbf{22M $\pm$ 0.27M}                                      & 28M $\pm$ 0.39K                                              &  &  &  &  &  &  &  &  &  \\
                                                                                  & NCPR = 200                   & \textbf{790 $\pm$ 16}                                    & 1481 $\pm$ 33                                           & \textbf{57M $\pm$ 1.0M}                                       & 59M $\pm$ 1.3M                                               &  &  &  &  &  &  &  &  &  \\ \cline{1-6}
                                                                                  &                              & \multicolumn{1}{l}{}                                  & \multicolumn{1}{l}{}                                 & \multicolumn{1}{l}{}                                       & \multicolumn{1}{l}{}                                      &  &  &  &  &  &  &  &  & 
\end{tabular}
\end{adjustbox}
\endgroup
\end{table}

\begin{toappendix}

This section summarizes the missing details from Section \ref{sec:sims}.  As mentioned, the datasets, models and tasks are exactly the same as the "EMNIST CR" task and the "SO NWP" task from \citet{reddi2020adaptive}Reddi el al [2020], such that we can use their optimized FedAvg results as baseline.  However, \citet{reddi2020adaptive}Reddi el al [2020] implement their algorithms on the Tensorflow Federated framework (\citet{tensorflowfederated}Ingerman et al. [2019]), whereas for our work, we build our algorithms on the FedJAX framework (\citet{ro2021fedjax}Ro et al. [2021]) which is under the Apache License.  
\subsection{Datasets, Models, and Tasks} \label{sec:app-sims}
We train a CNN to do character recognition (EMNIST CR) on the federated EMNIST-62 dataset (\citet{cohen2017emnist}Cohen et al. [2017]).  Next, we train a RNN to do next-word-prediction (SO NWP) on the federated Stack Overflow dataset (\citet{authors2019tensorflow_so}Authors [2019]).  

\paragraph{Federated EMNIST-62 with CNN}
EMNIST consists of images of digits and upper and lower case English characters, with 62 total classes. The federated version of EMNIST (\citet{caldas2018leaf}Caldas et al., 2018) partitions the digits by their author. The dataset has natural heterogeneity stemming from the writing style of each person.  See Table \ref{tbl:datastats} for more on the statistics of the federated EMNIST-62 dataset. On our select task of character recognition for this dataset (EMNIST CR),  a Convolutional Neural Network (CNN) is used.
The network has two convolutional layers (with $3 \times 3$ kernels), max pooling, and dropout, followed by a 128 unit dense
layer. A full description of the model is in Table \ref{tbl:emnist-cnn}.

\begin{table}
  \caption{EMNIST character recognition model architecture.}
  \label{tbl:emnist-cnn}
\begingroup
\setlength{\tabcolsep}{5pt} 
\renewcommand{\arraystretch}{1.2} 
  \centering
  \begin{tabular}{c c c c c} 
    \toprule
    \cmidrule(r){1-5}
    Layer & Output Shape & \# of Trainable Parameters & Activation & Hyperparameters \\
    \midrule
    \hline
    Input & $(28,28,1)$ & 0 & & \\
    Conv2d & $(26,26,32)$ & 320 & & kernel size = 3; strides = $(1,1)$ \\
    Conv2d & $(24,24,64)$ & 18496 & ReLU & kernel size = 3; strides = $(1,1)$ \\
    MaxPool2d & $(12,12,64)$ & 0 & & pool size = $(2,2)$ \\
    Dropout & $(12,12,64)$ & 0 & & p = 0.25 \\
    Flatten & 9216 & 0 & & \\
    Dense & 128 & 1179776 & & \\
    Dropout & 128 & 0 & & p = 0.5 \\
    Dense & 62 & 7998 & softmax & \\
    \bottomrule
  \end{tabular}
\endgroup
\end{table}

\paragraph{Federated Stack Overflow with RNN}
Stack Overflow is a language modeling dataset consisting of question and answers from the question and answer site, Stack Overflow. The questions and answers also have associated metadata, including tags. The dataset contains 342,477
unique users which we use as clients.  See Table \ref{tbl:datastats} for more on the statistics of the federated Stack Overflow dataset. We perform next-word prediction (Stack Overflow NWP, SO NWP for short) on this dataset. We restrict the task to the 10,000 most frequently used words, and each client to the first 128 sentences in their dataset. We also perform padding and truncation to ensure that sentences have 20 words. We then represent the sentence as a sequence of indices
corresponding to the 10,000 frequently used words, as well as indices representing padding, out-of-vocabulary words, beginning of sentence, and end of sentence. We perform next-word-prediction on these sequences using a Recurrent Neural Network (RNN) that
embeds each word in a sentence into a learned 96-dimensional space. It then feeds the embedded words into a single
LSTM layer of hidden dimension 670, followed by a densely connected softmax output layer. A full description of the model is in Table \ref{tbl:so-nwp}. The metric used in the main body is the top-1 accuracy over the proper 10,000-word vocabulary; that is, it does not include padding, out-of-vocab, or beginning or end of sentence tokens.

\begin{table}
  \caption{Stack Overflow next word prediction model architecture.}
  \label{tbl:so-nwp}
  \centering
\begingroup
\setlength{\tabcolsep}{10pt} 
\renewcommand{\arraystretch}{1.2} 
  \begin{tabular}{c c c} 
    \toprule
    \cmidrule(r){1-3}
    Layer & Output Shape & \# of Trainable Parameters  \\
    \midrule
    \hline
    Input & 20 & 0  \\
    Embedding & $(20, 96)$ & 960384 \\
    LSTM & $(20, 670)$ & 2055560 \\
    Dense & $(20,96)$ & 64416 \\
    Dense & $(20, 10004)$ & 970388 \\
    \bottomrule
  \end{tabular}
 \endgroup
\end{table}

\begin{table}
  \caption{Data statistics.}
  \label{tbl:datastats}
  \centering
\begingroup
\setlength{\tabcolsep}{10pt} 
\renewcommand{\arraystretch}{1.2} 
\begin{tabular}{c c c c c} 
    \toprule
    \cmidrule(r){1-5}
    Dataset & Train Clients & Train Examples & Test Clients & Test Examples \\
    \midrule
    \hline
    EMNIST-62 & 3,400 & 671,585 & 3,400 & 77,483 \\
    STACKOVERFLOW & 342,477 & 135,818,730 & 204,088 & 16,586,035 \\
    \bottomrule
  \end{tabular}
 \endgroup
\end{table}

\subsection{Client Sampling}
In all our experiments, we do not include updates from all clients in each communication round.  Instead, client sampling  is done, where clients are sampled uniformly at random from all training clients, without replacement within a given round, but with replacement across rounds.  In our EMNIST CR experiments, 10 out of a total of 3,400 clients are sampled in each communication round, and in our SO NWP experiments, 50 out of a total of 342,477 clients are sampled in each round.  

\subsection{Additional Results}

Below, we provide additional results from our experiments conducted in Section \ref{sec:sims} and whose test accuracy performance results shown in Figure \ref{fig:fig1}.  The baseline values were selected for best performance from \citet{reddi2020adaptive}Reddi et al. [2020].

\begin{figure} [H]
  \centering
  \makebox[0pt]{\includegraphics[scale=0.29]{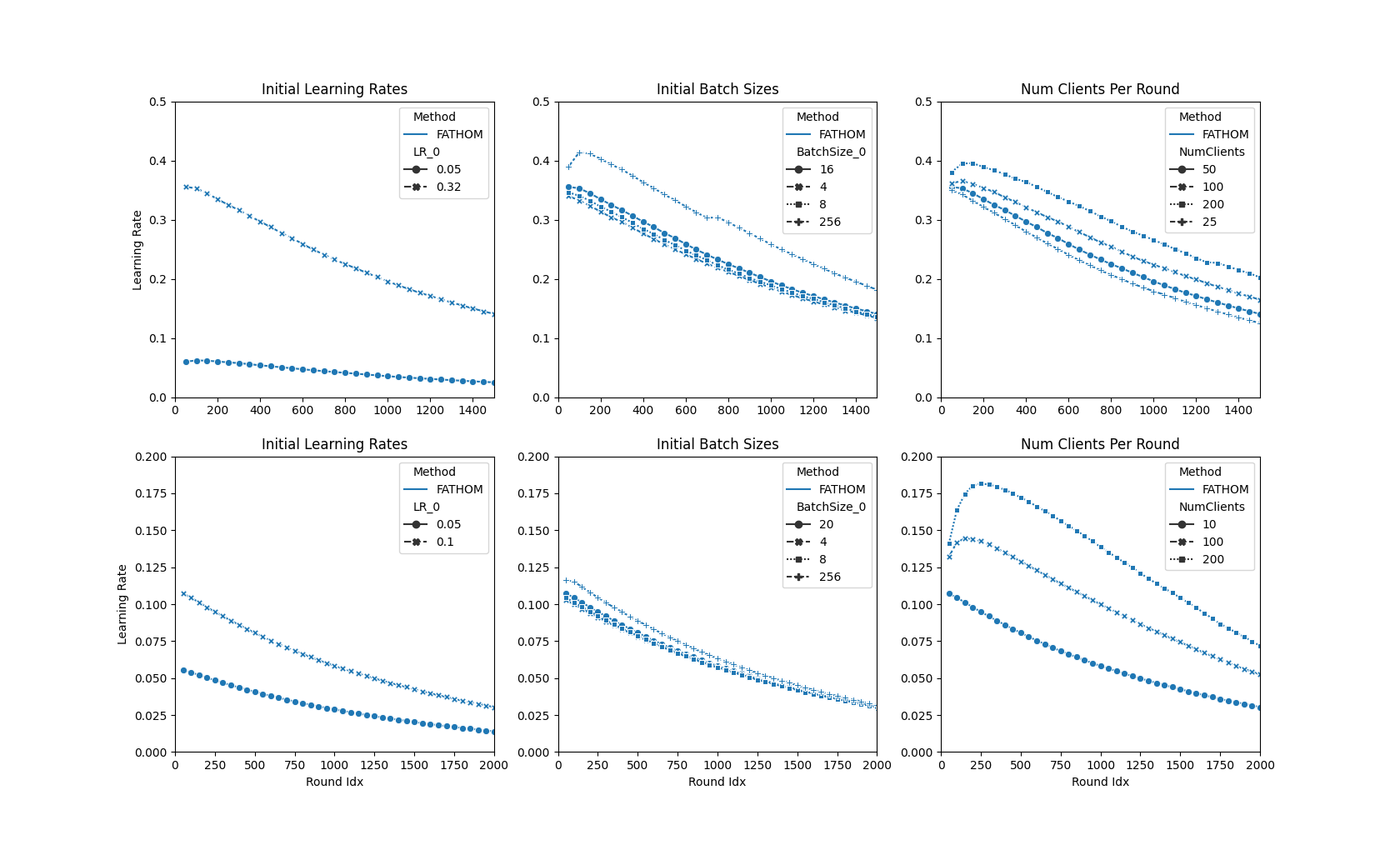}} 
  \caption{Adaptive client learning rate from the same experiments conducted in Section \ref{sec:sims} and in Figure \ref{fig:fig1}. Top row: FSO sims. Bottom row: FEMNIST sims.  Baseline values for FEMNIST: LR\_0=0.1, BatchSize\_0=20, NumClients=10.  Baseline values for FSO: LR\_0=0.32, BatchSize\_0=16, NumClients=50. } \label{fig:fig-LR}
\end{figure}

\begin{figure} [H]
  \centering
  \makebox[0pt]{\includegraphics[scale=0.29]{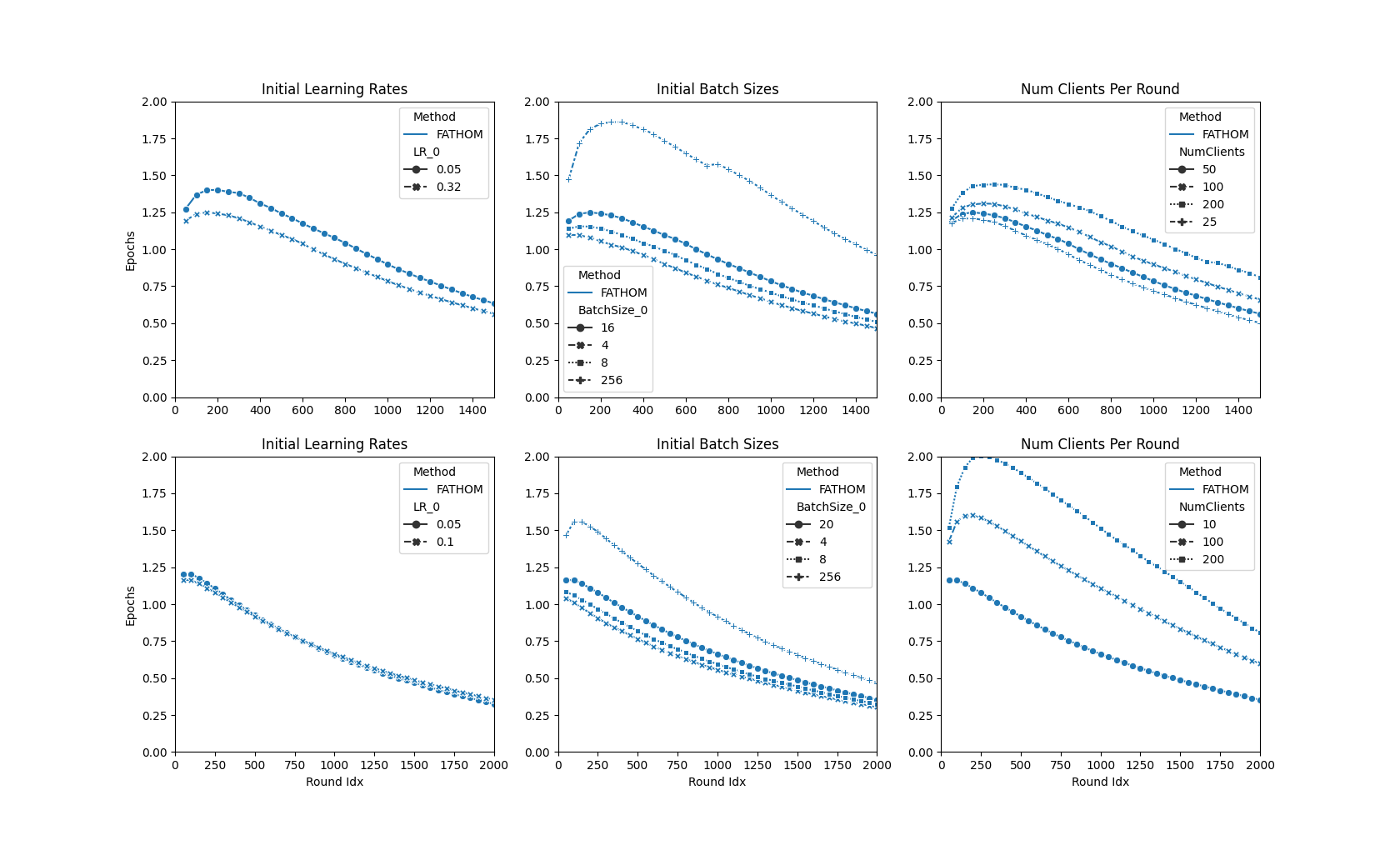}} 
  \caption{Adaptive number of epochs from the same experiments conducted in Section \ref{sec:sims}, and in Figures \ref{fig:fig1} and \ref{fig:fig-LR}. Top row: FSO sims. Bottom row: FEMNIST sims.  Baseline values for FEMNIST: LR\_0=0.1, BatchSize\_0=20, NumClients=10.  Baseline values for FSO: LR\_0=0.32, BatchSize\_0=16, NumClients=50. } \label{fig:fig-Ep}
\end{figure}

\begin{figure} [H]
  \centering
  \makebox[0pt]{\includegraphics[scale=0.29]{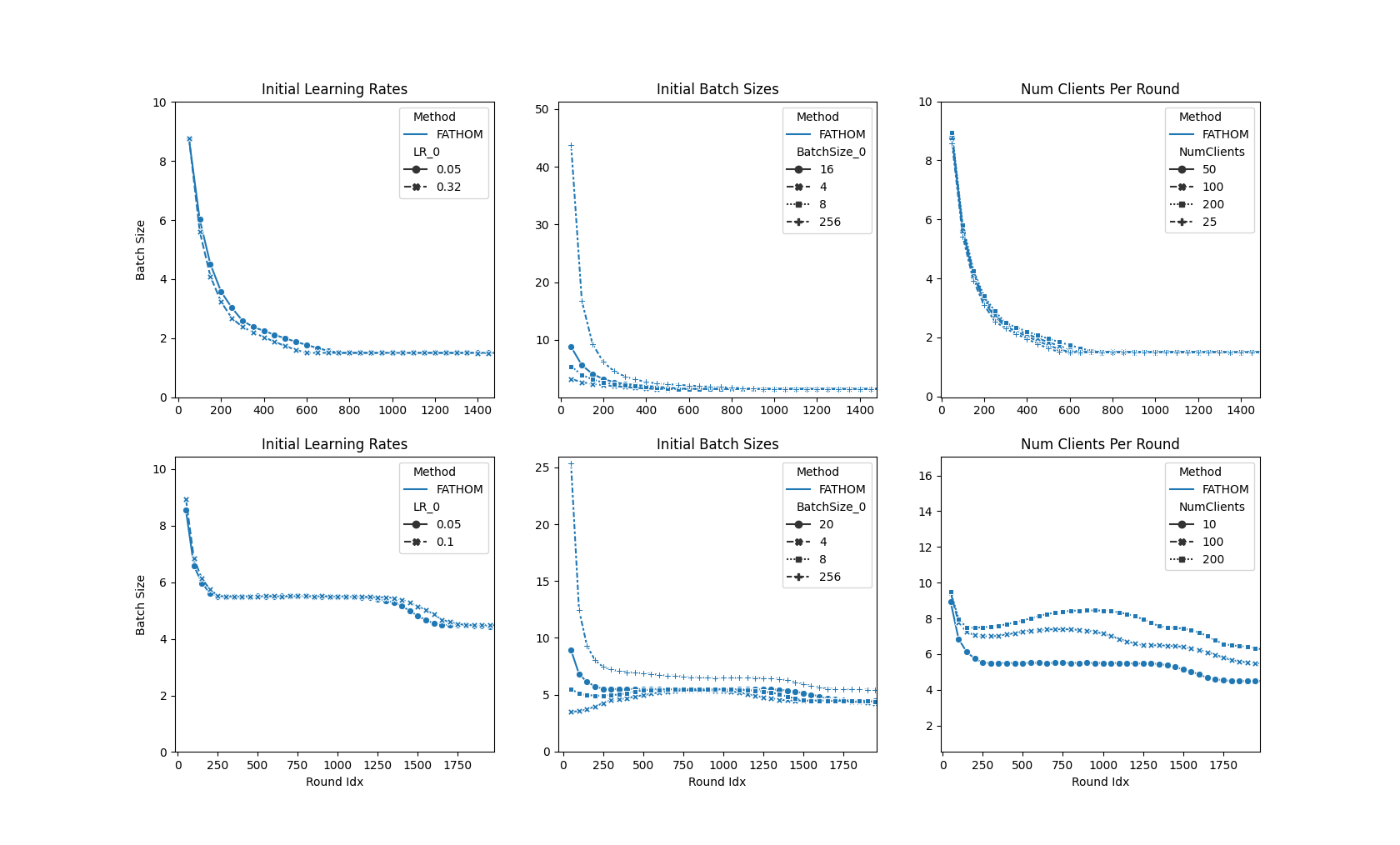}} 
  \caption{Adaptive batch size from the same experiments conducted in Section \ref{sec:sims}, and in Figures \ref{fig:fig1}, \ref{fig:fig-LR} and \ref{fig:fig-Ep}. Top row: FSO sims. Bottom row: FEMNIST sims.  Baseline values for FEMNIST: LR\_0=0.1, BatchSize\_0=20, NumClients=10.  Baseline values for FSO: LR\_0=0.32, BatchSize\_0=16, NumClients=50. } \label{fig:fig-BS}
\end{figure}

\end{toappendix}

\section{Conclusion and Future Work} \label{sec:conclusion}
In this work, we propose FATHOM for adaptive hyperparameters in federated optimization, specifically for FedAvg.  We analyze theoretically and evaluate empirically its potential benefits in convergence behavior as measured in test accuracy, and in reduction of local computations, by automatically adapting the three main hyperparameters of FedAvg: client learning rate, and number of local steps via epochs and batch size.  An example of future efforts to extend this work is using a standardized benchmark such as \citet{wang2022fedhpob} for performance comparison against other FL HPO methods.

\begin{ack}
This material is based on work supported by the Naval Information Warfare Center Pacific, specifically, funded by the Naval Innovative Science and Engineering (NISE) program which has been established by the Secretary of Defense, in consultation with the Secretaries of the Military to provide a mechanism for funding research and development within the laboratories of the Department of Defense.  Any opinions, findings and conclusions or recommendations expressed in this material are those of the author(s) and do not
necessarily reflect the views of any of these funding agencies.  
\end{ack}

\pagebreak

\medskip
{
\small
\bibliography{references}

\begin{thebibliography}{35}
\providecommand{\natexlab}[1]{#1}
\providecommand{\url}[1]{\texttt{#1}}
\expandafter\ifx\csname urlstyle\endcsname\relax
  \providecommand{\doi}[1]{doi: #1}\else
  \providecommand{\doi}{doi: \begingroup \urlstyle{rm}\Url}\fi

\bibitem[Amid et~al.(2022)Amid, Anil, Fifty, and Warmuth]{amid2022stepsize}
E.~Amid, R.~Anil, C.~Fifty, and M.~K. Warmuth.
\newblock Step-size adaptation using exponentiated gradient updates, 2022.

\bibitem[Baydin et~al.(2017)Baydin, Cornish, Rubio, Schmidt, and
  Wood]{baydin2017online}
A.~G. Baydin, R.~Cornish, D.~M. Rubio, M.~Schmidt, and F.~Wood.
\newblock Online learning rate adaptation with hypergradient descent, 2017.

\bibitem[Cai et~al.(2019)Cai, Zhu, and Han]{cai2018proxylessnas}
H.~Cai, L.~Zhu, and S.~Han.
\newblock Proxyless{NAS}: Direct neural architecture search on target task and
  hardware.
\newblock In \emph{International Conference on Learning Representations}, 2019.
\newblock URL \url{https://arxiv.org/pdf/1812.00332.pdf}.

\bibitem[Charles and Konečný(2020)]{charles2020outsized}
Z.~Charles and J.~Konečný.
\newblock On the outsized importance of learning rates in local update methods,
  2020.

\bibitem[Cohen et~al.(2017)Cohen, Afshar, Tapson, and
  Van~Schaik]{cohen2017emnist}
G.~Cohen, S.~Afshar, J.~Tapson, and A.~Van~Schaik.
\newblock Emnist: Extending mnist to handwritten letters.
\newblock In \emph{2017 international joint conference on neural networks
  (IJCNN)}, pages 2921--2926. IEEE, 2017.

\bibitem[Dai et~al.(2020)Dai, Low, and Jaillet]{dai2020federated}
Z.~Dai, K.~H. Low, and P.~Jaillet.
\newblock Federated bayesian optimization via thompson sampling, 2020.

\bibitem[Dai et~al.(2021)Dai, Low, and Jaillet]{dai2021differentially}
Z.~Dai, B.~K.~H. Low, and P.~Jaillet.
\newblock Differentially private federated bayesian optimization with
  distributed exploration, 2021.

\bibitem[Ghai et~al.(2019)Ghai, Hazan, and Singer]{ghai2019exponentiated}
U.~Ghai, E.~Hazan, and Y.~Singer.
\newblock Exponentiated gradient meets gradient descent, 2019.

\bibitem[Gorbunov et~al.(2020)Gorbunov, Hanzely, and
  Richtárik]{gorbunov2020local}
E.~Gorbunov, F.~Hanzely, and P.~Richtárik.
\newblock Local sgd: Unified theory and new efficient methods, 2020.

\bibitem[Guo et~al.(2022)Guo, Yang, Hatamizadeh, Xu, Xu, Li, Zhao, Xu, Harmon,
  Turkbey, et~al.]{guo2022auto}
P.~Guo, D.~Yang, A.~Hatamizadeh, A.~Xu, Z.~Xu, W.~Li, C.~Zhao, D.~Xu,
  S.~Harmon, E.~Turkbey, et~al.
\newblock Auto-fedrl: Federated hyperparameter optimization for
  multi-institutional medical image segmentation.
\newblock \emph{arXiv preprint arXiv:2203.06338}, 2022.

\bibitem[Hardy et~al.(1988)Hardy, Littlewood, and
  P{\'o}lya]{hardy1988inequalities}
G.~Hardy, J.~Littlewood, and G.~P{\'o}lya.
\newblock \emph{Inequalities}.
\newblock Cambridge Mathematical Library. Cambridge University Press, 1988.
\newblock ISBN 9781107647398.
\newblock URL \url{https://books.google.com/books?id=EfvZAQAAQBAJ}.

\bibitem[Holly et~al.(2021)Holly, Hiessl, Lakani, Schall, Heitzinger, and
  Kemnitz]{holly2021evaluation}
S.~Holly, T.~Hiessl, S.~R. Lakani, D.~Schall, C.~Heitzinger, and J.~Kemnitz.
\newblock Evaluation of hyperparameter-optimization approaches in an industrial
  federated learning system, 2021.

\bibitem[Khodak et~al.(2019)Khodak, Balcan, and Talwalkar]{khodak2019adaptive}
M.~Khodak, M.-F. Balcan, and A.~Talwalkar.
\newblock Adaptive gradient-based meta-learning methods, 2019.

\bibitem[Khodak et~al.(2021)Khodak, Tu, Li, Li, Balcan, Smith, and
  Talwalkar]{khodak2021federated}
M.~Khodak, R.~Tu, T.~Li, L.~Li, N.~Balcan, V.~Smith, and A.~Talwalkar.
\newblock Federated hyperparameter tuning: Challenges, baselines, and
  connections to weight-sharing.
\newblock In A.~Beygelzimer, Y.~Dauphin, P.~Liang, and J.~W. Vaughan, editors,
  \emph{Advances in Neural Information Processing Systems}, 2021.
\newblock URL \url{https://openreview.net/forum?id=p99rWde9fVJ}.

\bibitem[Kingma and Ba(2014)]{kingma2014adam}
D.~P. Kingma and J.~Ba.
\newblock Adam: A method for stochastic optimization, 2014.

\bibitem[Kingma and Welling(2013)]{kingma2013autoencoding}
D.~P. Kingma and M.~Welling.
\newblock Auto-encoding variational bayes, 2013.

\bibitem[Li et~al.(2020)Li, Khodak, Balcan, and Talwalkar]{li2020geometryaware}
L.~Li, M.~Khodak, M.-F. Balcan, and A.~Talwalkar.
\newblock Geometry-aware gradient algorithms for neural architecture search,
  2020.

\bibitem[Li et~al.(2019)Li, Huang, Yang, Wang, and Zhang]{li2019convergence}
X.~Li, K.~Huang, W.~Yang, S.~Wang, and Z.~Zhang.
\newblock On the convergence of fedavg on non-iid data, 2019.

\bibitem[Li and Cevher(2018)]{2018egmconv}
Y.-H. Li and V.~Cevher.
\newblock Convergence of the exponentiated gradient method with armijo line
  search.
\newblock \emph{Journal of Optimization Theory and Applications}, 181\penalty0
  (2):\penalty0 588–607, Dec 2018.
\newblock ISSN 1573-2878.
\newblock \doi{10.1007/s10957-018-1428-9}.
\newblock URL \url{http://dx.doi.org/10.1007/s10957-018-1428-9}.

\bibitem[McMahan et~al.(2016)McMahan, Moore, Ramage, Hampson, and
  y~Arcas]{mcmahan2016communicationefficient}
H.~B. McMahan, E.~Moore, D.~Ramage, S.~Hampson, and B.~A. y~Arcas.
\newblock Communication-efficient learning of deep networks from decentralized
  data, 2016.

\bibitem[Mokhtari et~al.(2016)Mokhtari, Shahrampour, Jadbabaie, and
  Ribeiro]{Mokhtari2016}
A.~Mokhtari, S.~Shahrampour, A.~Jadbabaie, and A.~Ribeiro.
\newblock Online optimization in dynamic environments: Improved regret rates
  for strongly convex problems.
\newblock \emph{2016 IEEE 55th Conference on Decision and Control (CDC)}, Dec
  2016.
\newblock \doi{10.1109/cdc.2016.7799379}.
\newblock URL \url{http://dx.doi.org/10.1109/cdc.2016.7799379}.

\bibitem[Mostafa(2019)]{mostafa2019robust}
H.~Mostafa.
\newblock Robust federated learning through representation matching and
  adaptive hyper-parameters, 2019.

\bibitem[Murota(1998)]{Murota1998DiscreteCA}
K.~Murota.
\newblock Discrete convex analysis.
\newblock \emph{Mathematical Programming}, 83:\penalty0 313--371, 1998.

\bibitem[Oppenheim and Schafer(2009)]{10.5555/1795494}
A.~V. Oppenheim and R.~W. Schafer.
\newblock \emph{Discrete-Time Signal Processing}.
\newblock Prentice Hall Press, USA, 3rd edition, 2009.
\newblock ISBN 0131988425.

\bibitem[Pham et~al.(2018)Pham, Guan, Zoph, Le, and Dean]{pmlr-v80-pham18a}
H.~Pham, M.~Guan, B.~Zoph, Q.~Le, and J.~Dean.
\newblock Efficient neural architecture search via parameters sharing.
\newblock In J.~Dy and A.~Krause, editors, \emph{Proceedings of the 35th
  International Conference on Machine Learning}, volume~80 of \emph{Proceedings
  of Machine Learning Research}, pages 4095--4104. PMLR, 10--15 Jul 2018.
\newblock URL \url{https://proceedings.mlr.press/v80/pham18a.html}.

\bibitem[Reddi et~al.(2020)Reddi, Charles, Zaheer, Garrett, Rush, Konečný,
  Kumar, and McMahan]{reddi2020adaptive}
S.~Reddi, Z.~Charles, M.~Zaheer, Z.~Garrett, K.~Rush, J.~Konečný, S.~Kumar,
  and H.~B. McMahan.
\newblock Adaptive federated optimization, 2020.

\bibitem[Ro et~al.(2021)Ro, Suresh, and Wu]{ro2021fedjax}
J.~H. Ro, A.~T. Suresh, and K.~Wu.
\newblock Fedjax: Federated learning simulation with jax.
\newblock \emph{arXiv preprint arXiv:2108.02117}, 2021.

\bibitem[TensorFlow-Federated-Authors(2019)]{authors2019tensorflow_so}
TensorFlow-Federated-Authors.
\newblock Tensorflow federated stack overflow dataset, 2019.
\newblock URL \url{https://www.tensorflow.
  org/federated/api\_docs/python/tff/simulation/datasets/stackoverflow}.

\bibitem[Wang and Joshi(2018)]{wang2018adaptive}
J.~Wang and G.~Joshi.
\newblock Adaptive communication strategies to achieve the best error-runtime
  trade-off in local-update sgd, 2018.

\bibitem[Wang et~al.(2021)Wang, Charles, Xu, Joshi, McMahan, y~Arcas,
  Al-Shedivat, Andrew, Avestimehr, Daly, Data, Diggavi, Eichner, Gadhikar,
  Garrett, Girgis, Hanzely, Hard, He, Horvath, Huo, Ingerman, Jaggi, Javidi,
  Kairouz, Kale, Karimireddy, Konecny, Koyejo, Li, Liu, Mohri, Qi, Reddi,
  Richtarik, Singhal, Smith, Soltanolkotabi, Song, Suresh, Stich, Talwalkar,
  Wang, Woodworth, Wu, Yu, Yuan, Zaheer, Zhang, Zhang, Zheng, Zhu, and
  Zhu]{wang2021field}
J.~Wang, Z.~Charles, Z.~Xu, G.~Joshi, H.~B. McMahan, B.~A. y~Arcas,
  M.~Al-Shedivat, G.~Andrew, S.~Avestimehr, K.~Daly, D.~Data, S.~Diggavi,
  H.~Eichner, A.~Gadhikar, Z.~Garrett, A.~M. Girgis, F.~Hanzely, A.~Hard,
  C.~He, S.~Horvath, Z.~Huo, A.~Ingerman, M.~Jaggi, T.~Javidi, P.~Kairouz,
  S.~Kale, S.~P. Karimireddy, J.~Konecny, S.~Koyejo, T.~Li, L.~Liu, M.~Mohri,
  H.~Qi, S.~J. Reddi, P.~Richtarik, K.~Singhal, V.~Smith, M.~Soltanolkotabi,
  W.~Song, A.~T. Suresh, S.~U. Stich, A.~Talwalkar, H.~Wang, B.~Woodworth,
  S.~Wu, F.~X. Yu, H.~Yuan, M.~Zaheer, M.~Zhang, T.~Zhang, C.~Zheng, C.~Zhu,
  and W.~Zhu.
\newblock A field guide to federated optimization, 2021.

\bibitem[Wang et~al.(2022)Wang, Kuang, Zhang, Ding, and Li]{wang2022fedhpob}
Z.~Wang, W.~Kuang, C.~Zhang, B.~Ding, and Y.~Li.
\newblock Fedhpo-b: A benchmark suite for federated hyperparameter
  optimization, 2022.

\bibitem[Williams(1992)]{10.1007/BF00992696}
R.~J. Williams.
\newblock Simple statistical gradient-following algorithms for connectionist
  reinforcement learning.
\newblock \emph{Mach. Learn.}, 8\penalty0 (3–4):\penalty0 229–256, may
  1992.
\newblock ISSN 0885-6125.
\newblock \doi{10.1007/BF00992696}.
\newblock URL \url{https://doi.org/10.1007/BF00992696}.

\bibitem[Yang et~al.(2021)Yang, Fang, and Liu]{yang2021achieving}
H.~Yang, M.~Fang, and J.~Liu.
\newblock Achieving linear speedup with partial worker participation in non-iid
  federated learning, 2021.

\bibitem[Zhou et~al.(2022)Zhou, Ram, Salonidis, Baracaldo, Samulowitz, and
  Ludwig]{zhou2022singleshot}
Y.~Zhou, P.~Ram, T.~Salonidis, N.~Baracaldo, H.~Samulowitz, and H.~Ludwig.
\newblock Single-shot hyper-parameter optimization for federated learning: A
  general algorithm and analysis, 2022.

\bibitem[Zinkevich(2003)]{10.5555/3041838.3041955}
M.~Zinkevich.
\newblock Online convex programming and generalized infinitesimal gradient
  ascent.
\newblock In \emph{Proceedings of the Twentieth International Conference on
  International Conference on Machine Learning}, ICML'03, page 928–935. AAAI
  Press, 2003.
\newblock ISBN 1577351894.

\end{thebibliography}
}

\appendix

\end{document}